\documentclass{article}

\usepackage[nonatbib,preprint]{neurips_2023}

\usepackage[utf8]{inputenc} 
\usepackage[T1]{fontenc}    
\usepackage{hyperref}       
\usepackage{url}            
\usepackage{booktabs}       
\usepackage{amsfonts}       
\usepackage{nicefrac}       
\usepackage{microtype}      
\usepackage{xcolor}         

\usepackage{pifont}
\newcommand{\cmark}{\ding{51}}%
\newcommand{\xmark}{\ding{55}}%

\newcommand{\datasetname}{{\textbf{M2Lingual}}}
\newcommand{\Evol}{\textit{Evol}}

\usepackage{adjustbox}
\usepackage{multirow}
\usepackage{enumitem}
\usepackage{cleveref}
\usepackage{array}
\usepackage{longtable}
\usepackage{tcolorbox}
\usepackage{booktabs}

\title{~\datasetname:
Enhancing Multilingual, Multi-Turn Instruction Alignment in Large Language Models}

\author{
 Rishabh Maheshwary$^{\S}$, \hspace{0.5mm} Vikas Yadav$^{\S}$, \hspace{0.5mm} Hoang Nguyen$^{\dag}$\thanks{Work done during internship at ServiceNow \\} 
 \\ \textbf{Khyati Mahajan$^{\S}$,}  \textbf{ Sathwik Tejaswi Madhusudhan$^{\S}$} \\
$^{\S}$ ServiceNow \\
$^{\dag}$ University of Illinois at Chicago \\
 \texttt{\{rishabh.maheshwary, vikas.yadav\}} @servicenow.com$^{\S}$
}

\begin{document}

\maketitle

\begin{abstract}
  Collecting instruction fine-tuning (IFT) data is a resource and time intensive task, especially in multilingual settings where finding proficient native speakers is challenging. Moreover, traditional data collection is prone to privacy risks, toxicity and lacks scalability. While fully synthetic datasets are a promising alternative, research on their use in multilingual domain is limited as existing approaches still rely on machine translation to improve multilingual performance. To bridge this gap we introduce~\datasetname, the~first \emph{fully synthetic}, \emph{multi-turn} multilingual dataset having $175K$ conversations across $70$ languages with a balanced mix of high, low and mid-resourced languages.~\datasetname~is constructed using a cost-efficient and scalable method that uses our \emph{novel two-step} \Evol~prompt taxonomy to transform a small set of human written instructions to complex and challenging conversations. Results across~\emph{three} model families, \emph{six} baseline datasets and evaluation spanning $31$ languages demonstrates the effectiveness of~\datasetname~over other datasets. We contribute the 2 step \Evol~taxonomy and the first fully synthetic, general and task-oriented, multi-turn, multilingual dataset built with \Evol~- \datasetname~\textcolor{blue}{\url{https://huggingface.co/datasets/ServiceNow-AI/M2Lingual}} - containing 175K total IFT pairs, covering 70 languages and 17+ NLP tasks.
\end{abstract}

\section{Introduction}

The recent success of large language models (LLMs)~\cite{GPT4paper, jiang2024mixtral, llamapaper, team2023gemini} can be largely attributed to the availability of large, diverse, and
high quality instruction fine-tuning (IFT) datasets~\cite{alpaca, chiang2023vicuna, xu2023wizardlm}. However, the majority of IFT datasets are in English with very limited coverage for other languages~\cite{zhang2023instruction}.

Existing multilingual IFT datasets can be divided into those that require human involvement and those that rely on machine translation (Table~\ref{tab:all-datasets}). The development of human-involved datasets is resource-heavy, often requiring native speakers, which introduces potential for annotator errors, uneven data distribution, and privacy and toxicity concerns~\cite{abay2019privacy,zhao2024WildChat}. These challenges lead to low-complexity conversations~\cite{xu2023wizardlm} as well. Machine-translated datasets offer less resource-intensive methods to create the data, but suffer from translation artifacts known as \emph{translationese}~\cite{bizzoni2020human, vanmassenhove2021machine} that fail to capture linguistic nuances~\cite{wang2023seaeval}. In conjunction with limited language coverage, overly simple instructions, and unbalanced NLP task representation, most multilingual datasets are not multi-turn, limiting the ability of models to engage beyond single utterances~\cite{wei2023polylm}.

\definecolor{darkgreen}{rgb}{0.0, 0.5, 0.0}
\begin{table*}[tbp!]
\centering
\small
\begin{adjustbox}{width=1\textwidth}
\begin{tabular}{@{}lccccccccc@{}}
\toprule
\multirow{2}{*}{Dataset} &
  \multirow{2}{*}{Size} &
  \multirow{2}{*}{\begin{tabular}[c]{@{}c@{}}Multi\\ turn?\end{tabular}} &
  \multirow{2}{*}{Langs} &
  \multicolumn{2}{c}{Resource Level} &
  \multirow{2}{*}{\begin{tabular}[c]{@{}c@{}}Task \\ specific?\end{tabular}} &
  \multirow{2}{*}{\begin{tabular}[c]{@{}c@{}}General \\ instructions?\end{tabular}} &
  \multirow{2}{*}{\begin{tabular}[c]{@{}c@{}}Translated \\ dataset?\end{tabular}} &
  \multirow{2}{*}{\begin{tabular}[c]{@{}c@{}}Fully \\ synthetic?\end{tabular}} \\ \cmidrule(lr){5-6}

              &                                                          &     &     & Low  & High &     &     &     &     \\ \midrule 
Aya Dataset~\cite{ayadataset} & 200K IR pairs & \textcolor{red}{\xmark} & 70  & 37 (1)  & 32   & \textcolor{red}{\xmark}  & \textcolor{darkgreen}{\cmark}  & \textcolor{darkgreen}{\xmark}  & \textcolor{red}{\xmark}  \\ 
MultiAlpaca~\cite{wei2023polylm}   & 132K IR pairs  & \textcolor{red}{\xmark}  & 11   & 0     & 11    & \textcolor{red}{\xmark} & \textcolor{darkgreen}{\cmark} & \textcolor{darkgreen}{\xmark} & \textcolor{darkgreen}{\cmark} \\
M-Alpaca~\cite{chen2023monolingual}   & 52K IR pairs  & \textcolor{red}{\xmark}  & 12   & 0     & 12    & \textcolor{red}{\xmark} & \textcolor{red}{\xmark} & \textcolor{red}{\cmark} & \textcolor{darkgreen}{\cmark} \\
Bactrian-X~\cite{li2023bactrian}    & 3.4M IR pairs & \textcolor{red}{\xmark}  & 52  & 15(1)   & 36   & \textcolor{red}{\xmark}  & \textcolor{darkgreen}{\cmark} & \textcolor{red}{\cmark} & \textcolor{darkgreen}{\cmark} \\ 
\midrule
OpenAssistant~\cite{kopf2024openassistant} & 10K convs & \textcolor{darkgreen}{\cmark} & 35  & 3    & 32   & \textcolor{red}{\xmark}  & \textcolor{red}{\xmark}  & \textcolor{darkgreen}{\xmark}  & \textcolor{red}{\xmark}  \\
ShareGPT~\cite{sharegpt}      & 94K convs                                                & \textcolor{darkgreen}{\cmark} & 45  & 4 (2)   & 39   & \textcolor{darkgreen}{\cmark} & \textcolor{red}{\xmark}  & \textcolor{darkgreen}{\xmark}  & \textcolor{red}{\xmark} \\ 
WildChat~\cite{zhao2024WildChat}     & 1.04M convs                                              & \textcolor{darkgreen}{\cmark} & 74  & 21 (3)   & 50   & \textcolor{red}{\xmark}  & \textcolor{red}{\xmark}  & \textcolor{darkgreen}{\xmark}  & \textcolor{red}{\xmark} \\ 
\midrule
\datasetname           & 182K convs                                                        & \textcolor{darkgreen}{\cmark}   & 70   & 37 (1)    & 32    &  \textcolor{darkgreen}{\cmark}   &  \textcolor{darkgreen}{\cmark}   &  \textcolor{darkgreen}{\xmark}   & \textcolor{darkgreen}{\cmark}   \\ \bottomrule
\end{tabular}
\end{adjustbox}
\caption{Comparison of multilingual IFT datasets with \datasetname. The top 4 rows are task based multilingual focused IFT datasets and the bottom 3 rows are datasets collected in the wild. Resource level classification taken from NLLB \cite{costa2022no}. Languages not found in the NLLB table are counted as low, in parentheses.}
\vspace{-3mm}
\label{tab:all-datasets}
\end{table*}


Fully synthetic datasets offer a promising solution to address the above concerns. Not only do synthetic datasets address the high cost of data collection, toxicity and privacy concerns, english synthetic datasets like WizardLM, Vicuna, Ultrachat, etc have been proven to significantly enhance the performance of LLMs in English~\cite{xu2023wizardlm,chiang2023vicuna,ding2023enhancing}. However, there is a lack of research on synthetic datasets in the multilingual domain that encompass a wide range of languages, NLP tasks, and multi-turn conversations.
To address this gap, we present the following contributions: 

\begin{enumerate}[noitemsep, nolistsep, left=0pt]

  \item We introduce \datasetname, the first \textit{fully synthetic}, \textit{multi-turn}, and \textit{diverse} multilingual dataset, containing $175K$ complex and challenging conversations across $70+$ languages and $19$ NLP tasks built with the \Evol~taxonomy.

  \item We construct a \textit{novel, two-step} \Evol~taxonomy (\Cref{fig:taxonomy}), covering $19$ NLP tasks, each with $9$ distinct methods to transform seed instructions to make them more complex and challenging. Additionally, to synthesize multi-turn conversations, we develop $21$~\Evol~prompts to increase engagement. This controlled setup \textit{ensures a balances representation of different languages}, especially low resource languages (Figure~\ref{fig:distribution}) which is challenging to achieve in real-world scenarios~\cite{przystupa2019neural}. The \Evol~taxonomy enables a fully-synthetic, scalable, and cost-efficient method for constructing enriched multi-turn multilingual conversational IFT dataset which is extendable to any task and language.

  \item We provide detailed analyses highlighting the impact of seed instructions, each step of the data enrichment and synthesis process. Additional analysis on low resource languages, content moderation, conversation length, and language distribution, demonstrate the superiority of \datasetname~over other datasets.
  
\end{enumerate}

\section{Related Work} \label{sec:related-work}

\paragraph{Multilingual Instruction Finetuning.} Due to the widespread availability of high-resource language pretraining corpora multilingual instruction finetuning has proven to be a cost effective solution for improving performance~\cite{ranaldi2023empowering,chen2023monolingual,ayamodel}. 
Several approaches have been adopted to expand access to multilingual IFT corpora. Notable among these are datasets derived from NLP tasks (e.g., FlanT5, Supernatural Instructions)~\cite{chung2024scaling, sanh2021multitask, wang2022super}
\begin{figure*}[htbp!]
  \centering
  \small
  \includegraphics[width=\linewidth, height=0.43\linewidth]{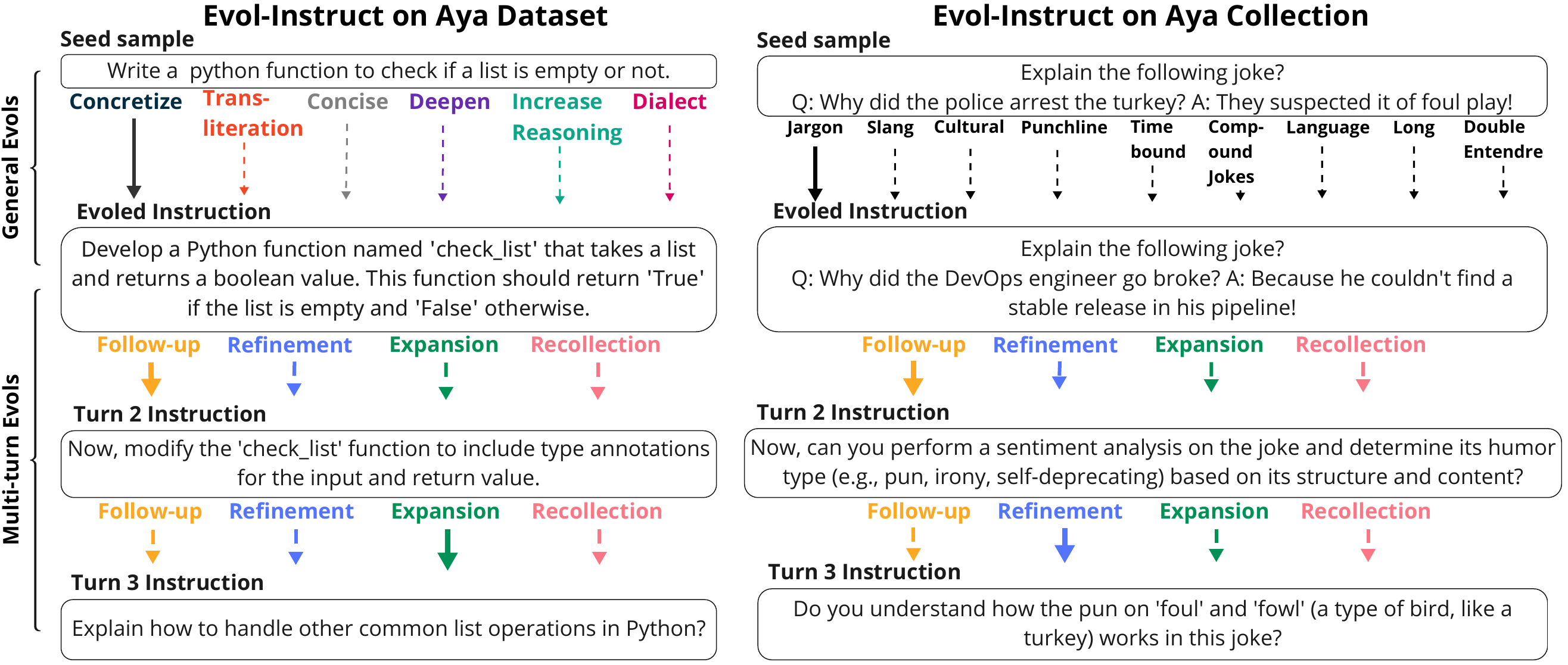}
  \caption{Walk-through for data synthesis of \datasetname. Step 1 is seed selection. In Step 2 for each instruction corresponding task specific~\Evol~prompt taxonomy is used for generating complex evoled instruction. Finally, in Step 3, multi-turn instruction are generated on Step 2 evoled instructions using multi-turn~\Evol~prompt taxonomy.}
  \label{fig:walkthrough-fig}
\end{figure*}
\emph{Human-generated datasets} such as Aya~\cite{ayadataset} and OpenAssistant \cite{kopf2024openassistant} involve humans creating conversation topics, writing questions, and crafting responses. While these datasets are typically high quality, their creation is extremely resource and time intensive. Moreover, finding native speakers for diverse languages is challenging, with potential annotator errors and uneven data distribution.~\cite{ayadataset, gilardi2023chatgpt}, making it difficult to scale these datasets. 
\emph{Human-AI generated datasets} such as LM-Sys~\cite{zheng2023LMSYS}, WildChat~\cite{zhao2024WildChat} and ShareGPT~\cite{sharegpt} are less resource-intensive than purely human-generated ones, as they involve humans interacting with LLMs to generate conversations. However, they still present challenges, as humans must write instructions and create diverse questions in native languages, a process that remains time-consuming. Additionally, this approach can raise privacy concerns~\cite{abay2019privacy}, introduce toxic data~\cite{zhao2024WildChat}, and result in low-complexity conversations~\cite{xu2023wizardlm}.
Finally, \emph{machine-translated datasets} such as BactrainX~\cite{li2023bactrian} offer a more resource efficient method. However, such datasets often suffer from translation artifacts known as \emph{translationese}~\cite{bizzoni2020human, vanmassenhove2021machine} that fail to capture linguistic nuances~\cite{wang2023seaeval}. On the contrary to these, our presented \datasetname~dataset utilizes IFT seeds from native speakers across various languages (\cref{sec:seed_selection_section}) and applies task specific mutation in each language (\cref{sec:task-guided-evols}), thus maintaining linguistic nuances in respective individual language. \datasetname's generation pipeline is also completely synthetic (\cref{tab:all-datasets}), making it a scalable and affordable for multilingual data generation.

\paragraph{Synthetic Datasets.} Fully synthetic datasets have emerged as a promising alternative towards addressing constraints with existing data generation methods. Popular English synthetic datasets, such as Alpaca~\cite{alpaca}, WizardLM~\cite{xu2023wizardlm}, and Vicuna~\cite{chiang2023vicuna}, generate new instructions from a small initial set using methods like Self-instruct~\cite{wang2023self} or \Evol-Instruct~\cite{xu2023wizardlm}, and have shown strong performance. However, there is limited research on leveraging synthetic datasets to enhance multilingual capabilities, with the exception of MultiAlpaca~\cite{wei2023polylm}, which uses Self-instruct. This approach has been shown to be susceptible to repetitive and noisy outputs~\cite{chenalpagasus,ghosh2024closer}, and suffers from low performance (Tables~\ref{tab:mt_bench_eval} \&~\ref{tab:evaluation}).
\section{Methodology} \label{sec:method}
\datasetname~has three main synthesis steps. \emph{Step 1: Seed Selection} involves the selection of diverse multilingual seeds. \emph{Step 2: Guided \Evol}~uses the \Evol~prompt taxonomy to generate complex instruction and response (IR) pairs and \emph{Step 3: Multiturn \Evol}~uses the multi-turn portion of the taxonomy to extend IR pairs to multilingual conversations. \Cref{fig:walkthrough-fig}~captures an overview of each step in \datasetname~synthesis and~\Cref{fig:taxonomy} presents the categories of \Evol~prompts\footnote{Complete \Evol~taxonomy prompts are in \Cref{sec:evol_prompts,sec:mt_evol_prompts}}.

\subsection{Seed Selection} \label{sec:seed_selection_section}
To ensure that we select diverse seeds capturing language nuances and covering a variety of NLP tasks, we select seed examples from the Aya dataset and collection as both have a high average approval ratio by human annotators~\cite{ayadataset}.

\noindent \textbf{Aya dataset seeds.} Aya dataset has general IR pairs written by native speakers that captures region specific language nuances and cultural contexts. We randomly select $100$ IR pairs for each of the $70$ languages, resulting in $7000$ seed IR pairs.

\noindent \textbf{Aya collection seeds.} Aya collection covers $19$ NLP tasks where each task has parallel examples in $113$ languages. To ensure a proper balance of the number of examples across all languages, we only focus on $70$ languages and exclude two NLP tasks --- 1) text simplification, as it is already supported by our \Evol~prompts, and 2) multilingual event entity task, as Aya do not have a consistent format for this task. Finally, for each task in the collection, we randomly sample $6$ examples per language, resulting in $6 \times 70 \times 17 = 7140$ IR seeds. We select $6$ random samples per task per language to ensure balanced amount of seed samples from Aya collection when compared to the seeds from Aya dataset. Thus, our final seeds contain $7000 + 7140 = 14140$ IR samples. 

\subsection{Guided \Evol} \label{sec:task-guided-evols}

\begin{figure*}[htbp!]
  \centering
  \small
  \includegraphics[width=\linewidth, height=0.5\linewidth]{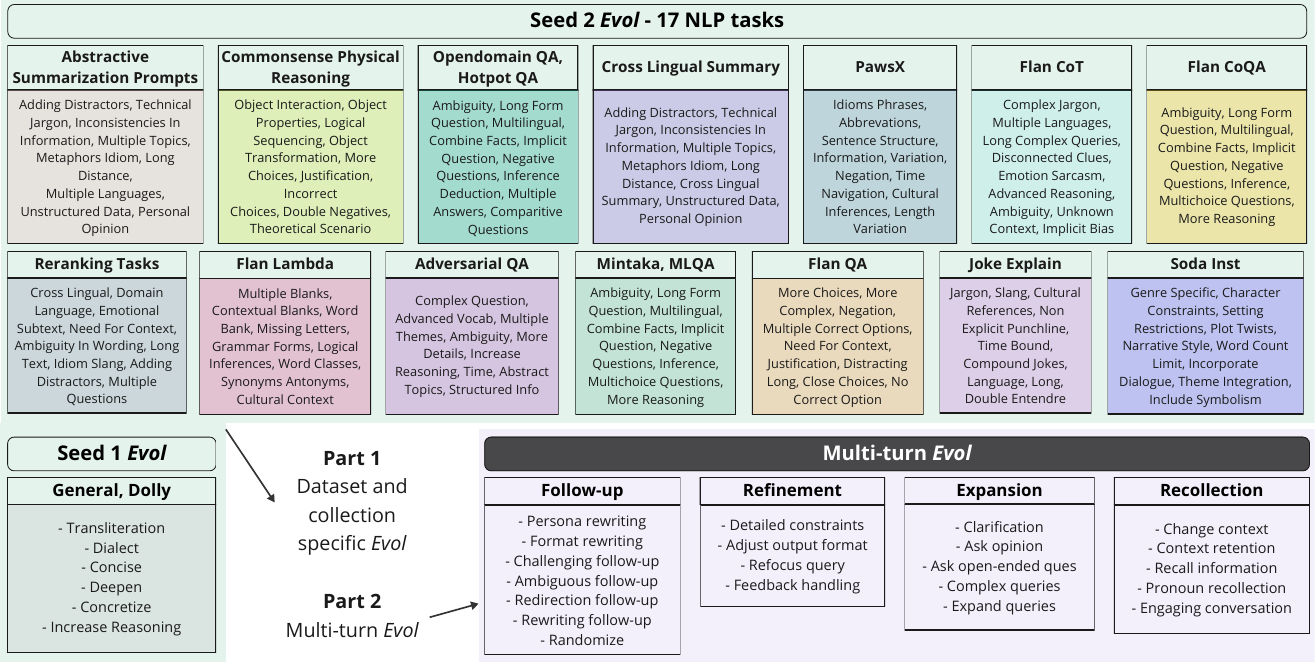}
  \caption{Taxonomy of \Evol~prompt conditions applied towards creating~\datasetname. Part 1 includes \Evol~prompts for Aya seeds and Part 2 has multi-turn \Evol~prompts applied for creating conversation.}
  \label{fig:taxonomy}
\end{figure*}

The seed instructions span a variety of NLP tasks but are generally straightforward and overly simplistic. To enhance LLMs' instruction following capabilities, particularly for complex tasks, we apply \textit{Evol-Instruct}~\cite{xu2023wizardlm} to our selected seed instructions. The \textit{Evol-Instruct} method uses Evol prompts to transform simple instructions into more intricate ones. However, the generic Evol conditions from the original work\footnote{\url{https://github.com/lcw99/evolve-instruct/blob/main/evolve.py}} provide very limited guidance for generating new IR pairs, especially for the $19$ diverse NLP tasks for which we aim to generate training data. To address this, we develop a novel \Evol~prompt taxonomy covering general instructions and NLP tasks, as shown in \Cref{fig:taxonomy}.

\noindent \textbf{\Evol~Prompt Taxonomy.} For general instructions from the Aya dataset, we design $6$ \Evol~prompt conditions that enhance multilingual features. We create $9$ task-specific~\Evol~prompts for each NLP task in the collection to ensure that we tailor \Evol~conditions for individual tasks. We use GPT-4~\cite{GPT4paper} to transform the seeds using our~\Evol~prompt taxonomy. These are applied to the seeds as follows:

\begin{itemize}[nolistsep, noitemsep, left=0pt]

  \item \emph{Aya dataset seeds.} As Aya dataset has general IR pairs, we apply the $6$ \textit{generic evol prompts} to each seed example. This results in $7K \times 6 = 42K$ instructions which are complex, challenging, and captures all nuances and complexities of languages.
  \item \emph{Aya collection seeds.} For each seed instruction from one of the $17$ tasks (top block of \Cref{fig:taxonomy}), we apply its corresponding $9$~\Evol s resulting in a total of $7140 \times 9 = 64260$ instructions.
\end{itemize}

\subsection{Multi-turn \Evol} \label{sec:multiturn-evols}
The final step involves generating multiple user-assistant turns based on the task-~\Evol ed instructions from the previous phase. Conversations between a user and an AI assistant generally fall into four broad categories: \textit{Follow-up}, \textit{Refinement}, \textit{Expansion}, and \textit{Recollection}~\cite{kwan2024mt}. We propose a multi-turn \Evol\ prompt taxonomy with $21$ distinct dialogue variations (final block labeled as part 2 in \Cref{fig:taxonomy}) that build upon the original generic four categories. Additionally, we ensure that all subsequent instructions are generated in the same language as the initial instruction by explicitly prompting GPT-4. We select all the \Evol ed instructions from the Aya dataset, and pick a balanced subset of size $35$K from Aya collection and generate turns as follows:

\begin{enumerate}[noitemsep, nolistsep, left=0pt]

  \item \emph{User turns.} We use the prompt specified in Appendix~\ref{sec:mt_evol_prompts} to generate multiple user turns. Specifically, we use the task-\Evol ed instruction from the previous step (i.e., Step 2), with its language and one of the $21$ dialogue variations to generate the next user instruction. 
  \item \emph{Assistant turns.} For all the generated user turns, we generate subsequent responses via GPT-4 using the entire conversation history. To mitigate the impact of topic drift from the long conversations \cite{zhang2021advances}, we restrict the total number of turns to $<=6$. 
\end{enumerate}

\paragraph{Post-Hoc Filtering.} Upon manual inspection, we find that some IR pairs generated using GPT-4 have repetitive long sequences and n-grams. To mitigate this, we apply a filtering step following~\cite{guo2021efficient, elmadany2023octopus} to remove IR pairs with frequent n-grams. This filtering is performed after steps $2$ and $3$. The final dataset consists of $75$K multi-turn conversations with $100$K single turn conversations. 



\section{Experiments}
We conduct experiments across \emph{three} model families \& \emph{five} model sizes --- Mistral-7B~\cite{jiang2023mistral}, LLaMA-3-8B~\cite{llamapaper} and QWEN-4B~\cite{bai2023qwen}. Furthermore, to demonstrate the effectiveness across different model scales, we fine-tune a larger model, LLaMA-2-13B~\cite{touvron2023llama}, and a smaller model, QWEN-1.8B~\cite{bai2023qwen}. To evaluate how well the datasets work with instruction-tuned models, we also experiment with Mistral-Instruct-7B.

\noindent \textbf{Baselines} --- We use 3 multilingual instruction finetuning (IFT) datasets \emph{MultiAlpaca}, \emph{Bactrian-X}, and \emph{Aya} for main evaluation. Furthermore, to highlight the importance of each step in our synthesis, we consider several ablations. Specifically, we train models using 1) only \textbf{Seed} samples, 2) seed samples with the generated~\Evol~s (\textbf{Seed + Evol}) and 3) seeds,~\Evol~s and the multi-turn conversations (\textbf{Seed + Evol + MT}). 
\subsection{Evaluation}
\paragraph{Multilingual benchmarks.} We utilize the EleutherAI evaluation \cite{eval-harness} for consistent comparisons on the following tasks:
  \begin{itemize}[noitemsep, nolistsep, left=0pt]
    \item \textit{Question Answering}: We focus on $3$ QA datasets 1) XQUAD~\cite{artetxe2019cross}, TyDiQA~\cite{clark2020tydi} and MLQA~\cite{lewis2019mlqa}. We use $3$ in-context examples and in the interest of time, we keep the number of examples per language to $100$ for XQUAD and MLQA, and $1000$ for TyDiQA. We use the validation set for XQUAD and test set for TyDiQA \& MLQA with F1-score as the metric.

    \item \textit{Summarization}: We use XLSUM~\cite{hasan2021xl} on $6$ languages --- Arabic, English, Spanish, French, Japanese and Russian with $100$ examples per language and use GPT-4 as a judge to rate the generated summaries on a scale of 1 to 5. For comparison, we also report ROUGE\textsubscript{L}~\cite{lin-2004-rouge} \& BLEU~\cite{papineni-etal-2002-bleu}.
    \item \textit{Multilingual math word problems}: We use MGSM~\cite{shi2022language} that translates GSM8K~\cite{cobbe2021training} to $10$ languages. We use $3$ in-context examples and compute exact match (EM) with ground truth answer.
\item \textit{Classification}: We focus on XNLI~\cite{conneau2018xnli} and XCOPA~\cite{ponti2020xcopa} with $15$ and $11$ languages respectively in a zero-shot setting and report the resuls in Appendix. We compute the accuracy (Acc) by looking at the log-likelihood assigned to the ground truth answer on the validation set.
  \end{itemize} 

\paragraph{Multilingual MT-Bench.} We evaluate conversational complex instruction following ability using MT-Bench~\cite{zheng2023judging}. 
It has $80$ multi-turn questions across $8$ domains. The models are required to respond to an initial and a follow-up question and GPT-4 assesses the responses on a scale of $1$ to $10$, with the overall score being the mean over two turns. We translate it into $8$ different languages (French, Canadian French, German, Italian, Spanish Japanese, Dutch, Portuguese) with professional linguists to ensure high quality evaluation. We modify the judge prompt to include the language of the question, and instruct GPT-4 to make sure the responses are in the same language. We report the average scores across $80$ examples for each language and the average score across all languages.

\noindent \textbf{Low-resource Evaluation.} To demonstrate the wide coverage of low resource languages in~\datasetname, we further evaluate models by on by translating MT-Bench to $6$ low-resource languages namely Hindi, Urdu, Thai, Tamil, Bengali and Gujarati using GPT-4. Finally, we also perform low-resource evaluation across 10 languages from Flores200~\cite{costa2022no, goyal2022flores}. We present BLEU~\cite{papineni-etal-2002-bleu} scores for translating each language into every other language. The final score for a language is the average BLEU score across all its translations to the remaining languages. The languages we used are Arabic (arb), Assamese (asm), Awadhi (awa), Belarusian (bel), Haitian Creole (hat), Kirghiz (kir), Burmese (mya), Nepali (nep), Somali (som), and Yoruba (yor). This selection covers a wide range of geographic regions (South Asia, Africa, Eastern Europe, and the Middle East) and includes languages with different writing systems: Latin, Cyrillic, Arabic, and Devanagari scripts.

\definecolor{darkgreen}{HTML}{b8d8b0}
\definecolor{lightgreen}{HTML}{c5eace}
\begin{table*}[htbp!]
  \centering
  \small
  \begin{adjustbox}{max width=\textwidth}
  \begin{tabular}{@{}c@{}ccccccc@{}}
    \toprule
    \multirow{2}{*}{\textbf{Model}} & \multirow{2}{*}{\textbf{Dataset}} 
    & \textbf{MT-AVG} & \textbf{XQUAD} & \textbf{TyDiQA} & \textbf{MLQA} & {\textbf{XLSUM}} & \textbf{MGSM} \\
     &  & & F1 & F1 & F1 & GPT-4 score & EM \\
    \midrule
    \multirow{6}{*}{\rotatebox[origin=c]{90}{\textbf{Mistral-7B}}}
    & MultiAlpaca & 4.77 & 67.99 & 64.44  & 55.69  & 3.06 & 11.71 \\
    & Bactrian-X & 5.25 & 71.91 & 66.63 & 60.27  & 2.59 & 17.14  \\
    & Aya & 5.18 & 70.46 & 66.95 & 57.47  & 3.13 & 13.86  \\
    \cmidrule[0.1pt](lr){2-8}
    & Seed & 5.04 & 72.52 & 65.89 & 59.33  & 2.94 & 13.65  \\
    & Seed + Evol & 5.56 & 71.01 & 65.04 & 57.47  & 3.41 & 18.38  \\
    & Seed + Evol + MT (\datasetname) &  \textbf{6.54} &  \textbf{74.53} &  \textbf{67.57} &  \textbf{62.40}  &  \textbf{3.55} &  \textbf{22.00}  \\
    \midrule
    \multirow{6}{*}{\rotatebox[origin=c]{90}{\textbf{LLaMA-3-8B}}}
    & MultiAlpaca & 4.41 & 75.08 & 64.49 &  59.01  & 3.17 & 10.68  \\
    & Bactrian-X & 5.66 & 69.57 & 56.45 &  58.51  & 2.79 & 22.86 \\
    & Aya & 4.95 & 75.14 & 59.60 & 53.14 &  3.29 & 22.09  \\
    \cmidrule[0.1pt](lr){2-8}
    & Seed & 3.54 &  \textbf{77.27} & 68.57 & 60.01  & 2.84 & 11.71  \\
    & Seed + Evol & 6.12 & 76.17 &  \textbf{69.89} & 63.09  & 3.52 &  \textbf{28.00}  \\
    & Seed + Evol + MT (\datasetname) &  \textbf{6.74} & 75.91 & 67.84 &  \textbf{63.5}  &  \textbf{3.68} & {27.36}  \\
    \bottomrule
  \end{tabular}
  \end{adjustbox}
  \caption{Performance comparison of Mistral-7B and LLaMA-3-8B. MT-Avg is average MT bench results across 9 languages (French, Canadian French, German, Italian, Spanish, Japanese, Dutch, Portuguese). \textit{Seeds} are $15.1$K seeds; \textit{Seed + Evol} is additional \Evol~IR pairs. \textit{Seed + Evol + MT} has additional multi-turn data.}
  \label{tab:combined_evaluation}
\end{table*}

\section{Results}

\paragraph{Multilingual MT-Bench}--- \Cref{tab:combined_evaluation} shows the average scores on MT-Bench across 9 languages. 
On average, \datasetname~outperforms other baseline datasets by $1.01$ MT-bench score with Mistral-7B, and $1.2$ with LLama-3-8B. The significant improvements across most baselines highlight the strengths of \datasetname. 
Detailed results across all 9 languages of MT-bench with different base models are shown in Appendix in Table~\ref{tab:mt_bench_eval} and~\ref{tab:mt_bench_eval_instruct}.

\paragraph{Multilingual NLP results}---
\datasetname~leads in performance on $3$ multilingual NLP tasks across Mistral-7B and LLama-3-8B. Specifically in Table~\ref{tab:combined_evaluation}, on average~\datasetname~always outperforms on QA task by ~$6-8\%$ and on MGSM by $8\%$ across both models. On summarization across both models our dataset outperforms by $0.5$ GPT-4 score on an average. We observed GPT-4 score to be a more reliable metric than BleU or ROUGEL score for evaluating summarization quality, as those metrics tend to un-fairly penalize long form LLM answers (more in \Cref{label:tok}). However for completeness we report results on BleU and ROUGEL in ~\Cref{tab:evaluation}. Finally, Table~\ref{tab:evaluation} and ~\ref{tab:evaluation_instruct} in Appendix has results on XNLI, XCOPA and other base models \Cref{sec:qweb4b}.
\paragraph{Importance of~\Evol}--- The \textbf{Seed + \Evol} rows in Table~\ref{tab:combined_evaluation} show the performance of seeds and synthetic~\Evol~IR pairs. In comparison to \textbf{Seed} data across both models, the average MT-Bench score increases by at least 1.75 points, with gains in every language (see Table~\ref{tab:mt_bench_eval} and~\ref{tab:mt_bench_eval_instruct}), especially Japanese. Similarly, \datasetname~leads to improvements of around $5$ and $10$ points on math word problem across Mistral-7B and LLama-3-8B respectively. On summarization and QA, the results are mixed (either increase, slight drop or same) as shown in Table~\ref{tab:combined_evaluation}. This is due to the increased verbosity of LLMs trained on \datasetname, which contains detailed synthetic~\Evol s. (more in \Cref{label:tok}). Finally, to demonstrate that the importance lies in the use of synthetic~\Evol s rather than simply increasing the amount of seed-like data, we sampled an additional $86K$ IR pairs from the Aya dataset and collection and replace it with synthetic generated~\Evol s and observed that the performance decreases across all benchmarks, especially in MT-Bench and MGSM by $1.1$ and $3.38$ points respectively (results are shown in \Cref{sec:evol_seed}).

\paragraph{Importance of multi-turn~\Evol.} The \textbf{Seed + \Evol~+ MT} rows in Table~\ref{tab:combined_evaluation}  shows performance after adding synthetically generated turns using multi-turn~\Evol. This boosts performance in MT-Bench evaluations substantially by $0.8$ points with the most significant gain of $1.31$ and $1.0$ points on French and Japanese for Mistral-7B model (Table~\ref{tab:mt_bench_eval}). Adding multi-turn data also helps in multilingual benchmarks as the results consistently improve by $3-4\%$ across all evaluations. Additional results are shown in Appendix, Table~\ref{tab:mt_bench_eval},~\ref{tab:evaluation}, ~\ref{tab:evaluation_instruct}, and, ~\ref{tab:mt_bench_eval_instruct}.

\section{Additional Analysis}

\definecolor{darkgreen}{HTML}{b8d8b0} 
\definecolor{lightgreen}{HTML}{c5eace}
\begin{table*}[h!]
\centering
\small
\begin{adjustbox}{max width=\textwidth}
\begin{tabular}{@{}c@{}ccccccc@{}}
\toprule
\multirow{2}{*}{\textbf{Model}} & \multirow{2}{*}{\textbf{Dataset}} & \textbf{MT-AVG} & \textbf{XQUAD} & \textbf{TyDiQA} & \textbf{MLQA} & {\textbf{XLSUM}} & \textbf{MGSM} \\
& & & F1 & F1 & F1 & GPT-4 score & EM \\
\midrule
\multirow{4}{*}{{\textbf{Mistral-7B}}} & Open Assistant & 5.66 & 67.99 & 54.22 & 53.64 & 2.60 & 16.05 \\
& ShareGPT & 5.80 & 66.33 & 56.97 & 50.78 & 2.41 & 11.32 \\
& WildChat & 6.53 & 72.55 & 64.27 & 59.53 & 2.93 & 18.41 \\
& \datasetname & \textbf{6.54} & \textbf{74.53} & \textbf{67.57} & \textbf{67.57} & \textbf{3.55} & \textbf{22.00} \\
\midrule
\multirow{4}{*}{{\textbf{LLaMA-3-8B}}} & Open Assistant & 5.12 & 64.38 & 52.65 & 47.08 & 2.92 & 17.36 \\
& ShareGPT & 6.10 & 56.98 & 58.48 & 43.43 & 2.54 & 25.32 \\
& WildChat & \textbf{6.75} & 63.15 & 59.88 & 63.16 & 2.90 & 26.36 \\
& \datasetname & 6.74 & \textbf{75.91} & \textbf{67.84} & \textbf{63.50} & \textbf{3.68} & \textbf{27.36} \\
\bottomrule
\end{tabular}
\end{adjustbox}
\caption{Results on Mistral-7B and LLaMA-3-8B with OpenAssitant, ShareGPT \& WildChat.}
\label{tab:combined_evaluation_data_in_wild}
\end{table*}

\begin{table*}[htbp!]
    \centering
    \small
    \begin{tabular}{@{}cccccccc@{}}
        \toprule
        \textbf{Model}  & \textbf{Dataset}  & \textbf{MT\_bn} & \textbf{MT\_gu} & \textbf{MT\_hi} & \textbf{MT\_ur} & \textbf{MT\_th} & \textbf{MT\_ta} \\
        \midrule
        \multirow{5}{*}{\textbf{Mistral-7B}} 
                        & MultiAlpaca    & 1.54      & 1.28       & 2.38     & 1.60    & 2.73    & 1.38    \\ 
                        & Bactrian-X        & 3.58       & 2.75      & 3.93     & 3.21    & 4.01    & 2.37    \\ 
                        & Aya        & 2.46       & 1.42        & 2.44     & 2.27    & 2.15    & 2.01    \\ 
                        & Wildchat    & 2.23       & 1.22        & 3.37     & 2.34   & 3.28    & 1.53    \\ 
                        & ~\datasetname   & \textbf{3.92}       & ~\textbf{3.3}         & ~\textbf{4.52}     & ~\textbf{3.80}    & ~\textbf{4.01}    & ~\textbf{2.57}    \\ 
        \midrule
        \multirow{5}{*}{\textbf{LLaMA-3-8B}} 
                        & MultiAlpaca      & 2.68      & 2.35        & 3.08     & 2.33    & 3.25    & 2.09    \\ 
                        & Bactrian-X        & 3.13       & 2.75      & 4.13     & 2.63    & 3.85    & 2.13    \\ 
                        & Aya         & 3.3        & 2.4         & 3.9      & 2.86    & 3.57    & 2.68    \\ 
                        & Wildchat   & 4.52       & 3.6         & 5.44     & 4.27    & 5.30    & 4.16    \\ 
                        & ~\datasetname & \textbf{4.73}       & \textbf{3.91}        & ~\textbf{5.97}     & ~\textbf{4.5}    & ~\textbf{5.68}    & ~\textbf{4.25}    \\ 
        \bottomrule
    \end{tabular}
    \caption{Low-resource evaluation of Aya, WildChat, and~\datasetname~using Mistral-7B and LLama-3-8B base models on Bengali (bn), Gujarati (gu), Hindi (hi), Urdu (ur), Thai (th), and Tamil (ta).}
    \label{tab:low_resource}
\end{table*}

\begin{table*}[ht]
    \centering
    \small
    \begin{tabular}{ccccccccccccc}
        \toprule
        \textbf{Model} & \textbf{Dataset} & \textbf{arb} & \textbf{asm} & \textbf{awa} & \textbf{bel} & \textbf{hat} & \textbf{kir} & \textbf{mya} & \textbf{nep} & \textbf{som} & \textbf{yor} & \textbf{Avg} \\
        \midrule
        \multirow{5}{*}{\textbf{Mistral-7B}} 
                                    & Multialpaca & 0.72 & 0.47 & 0.53 & 0.46 & 0.66 & 0.41 & 0.37 & 0.59 & 0.89 & 0.79 & 0.58 \\
                                    & Bactrian-X & 0.57 & 0.6  & 0.96 & 0.68 & 0.6  & 0.52 & 0.39 & 0.78 & 0.63 & 0.45 & 0.61 \\
                                    & Aya        & 0.86 & 0.52 & 0.64 & 0.74 & 1.42 & 0.61 & 0.41 & 0.57 & 0.83 & \textbf{1.38} & 0.79 \\
                                    & WildChat   & 0.5  & 0.48 & 0.66 & 0.62 & 0.97 & 0.76 & 0.47 & 0.67 & 0.68 & 0.76 & 0.65 \\
                                    & ~\datasetname~ & \textbf{1.31} & \textbf{0.94} & \textbf{0.91} & \textbf{0.82} & \textbf{1.57} & \textbf{0.83} & \textbf{0.56} & \textbf{1.2} & \textbf{1.05} & {1.01} & \textbf{1.02} \\
        \midrule
        \multirow{5}{*}{\textbf{LLama-3-8B}} 
                                    & Multialpaca & 1.74 & 0.88 & 1.04 & 0.84 & 1.29 & 1.02 & 0.74 & 1.34 & 0.96 & 1.2 & 1.1 \\
                                    & Bactrian-X & 1.51 & 0.89 & 1.02 & 1.13 & 1.23 & 1.00 & 0.41 & 1.09 & 1.13 & 0.88 & 1.02 \\
                                    & Aya        & 2.07 & 1.11 & 2.3  & 1.49 & 1.69 & 1.23 & 0.96 & 1.12 & \textbf{1.66} & 1.55 & 1.5 \\ 
                                    & WildChat   & 2.11 & 1.61 & \textbf{2.27} & 1.18 & 2.12 & 1.24 & 1.13 & 1.45 & 1.12 & 1.31 & 1.5 \\
                                    & ~\datasetname~ & \textbf{2.9} & \textbf{1.93} & 1.99 & \textbf{2.15} & \textbf{3.27} & \textbf{1.76} & \textbf{1.35} & \textbf{2.15} & 1.37 & \textbf{1.77} & \textbf{2.06} \\
        \bottomrule
    \end{tabular}
    \caption{Low-resource evaluation across 10 languages from Flores200. We present BLEU scores for translating each language into every other language. The final score for a language is calculated as the average BLEU score across all its translations to the remaining languages.}
    \label{tab:flores200_eval}
\end{table*}

\begin{table*}[htbp!]
\centering
\small
\begin{tabular}{@{}cccccccc@{}}
\toprule
\multirow{2}{*}{\textbf{Model}} & \multirow{2}{*}{\textbf{Dataset}} & \textbf{MT-AVG} & \textbf{XQUAD} & \textbf{TyDiQA} & \textbf{MLQA} & {\textbf{XLSUM}} & \textbf{MGSM} \\
& & & F1 & F1 & F1 & GPT-4 score & EM \\
\midrule
\multirow{3}{*}{Qwen-1.8B} & MultiAlpaca &  2.03 & 31.60 & 33.38 & 19.30 & 2.03 & 7.45 \\
              & WildChat  &  2.59 & 45.12 & 42.42 & 29.39 & 1.90 & 8.00 \\
              & \datasetname~&  \textbf{4.27} & \textbf{52.12} & \textbf{47.66} & \textbf{38.24} & \textbf{2.83} & \textbf{12.23} \\
\midrule
\multirow{3}{*}{LLaMA-2-13B} & MultiAlpaca &  4.46 & 55.07 & 59.46 & 48.74 & 2.48 & 7.80 \\
              & WildChat   &  6.00 & 67.64 & 60.14 & 53.69 & 2.55 & 9.95 \\
              & \datasetname~& \textbf{6.08} & \textbf{69.38}  & \textbf{64.66} & \textbf{54.64} & \textbf{3.03} & \textbf{11.95} \\
\bottomrule
\end{tabular}
\vspace{2mm}
\caption{Evaluations of QWEN-1.8B and LLaMa-2-13B for highlighting impact on different sized LLMs.}
\label{tab:different_sizedLLM_eval}
\end{table*}

\paragraph{Comparison with Human-AI generated data in the wild.} For completeness, we also include performance comparisons with Human-AI generated datasets collected from voluntary participation, such as OpenAssistant, ShareGPT, and WildChat in \Cref{tab:combined_evaluation_data_in_wild}, where \datasetname~shows strong performance results with both Mistral-7B and Llama-3-8B. Concretely,~\datasetname~outperforms OpenAssistant and ShareGPT by $0.8$ and $0.6$ on multilingual MT-Bench, $8\%$ and $12\%$ on QA, $0.7$ and $1.0$ on summarization and $8\%$ and $5\%$ on math word problem solving.~\datasetname~ performs comparable to WildChat on multilingual MT-Bench however strongly outperforms by $2-3\%$ on QA, $0.6$ on summarization and $2\%$ on math word problem solving. We report per language MT bench scores, results on other metrics and classification benchmarks in Appendix (Tables~\ref{tab:mt_bench_eval},~\ref{tab:evaluation}~\ref{tab:evaluation_instruct} and~\ref{tab:mt_bench_eval_instruct}). Finally we also compare the performance of~\datasetname~on low resource languages (see below) and observed that~\datasetname~notably outperforms these Human-AI generated datasets as shown in \cref{tab:low_resource} and \cref{tab:flores200_eval}. However, we would like to point out that these datasets are not focused specifically towards improving multilingual abilities, even though their creation methods lend inadvertently to multilingual data. For holistic comparison, we also include details with the next best performing dataset WildChat in further analysis.

\begin{figure*}[h!]
  \small
  \centering
  \includegraphics[width=\linewidth, height=.23\linewidth]{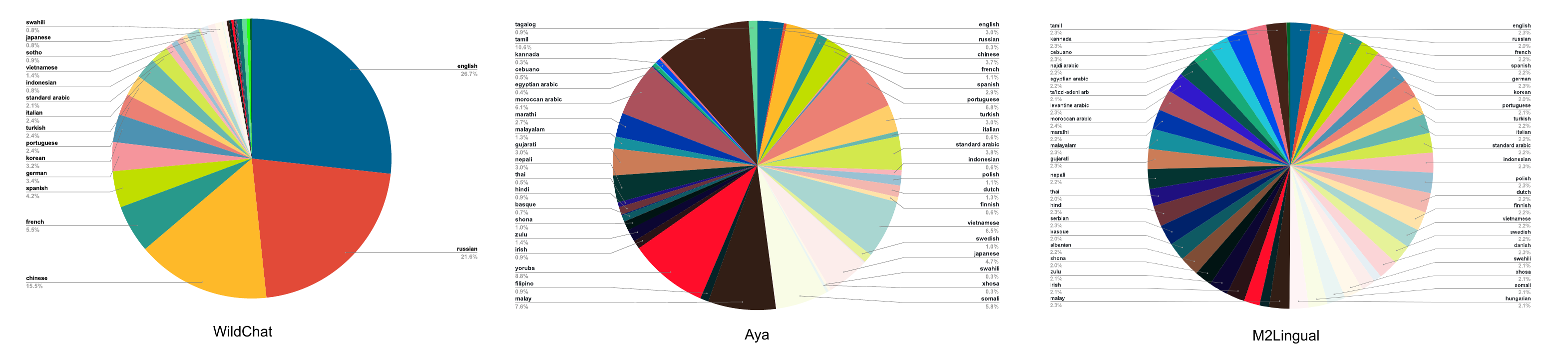}
  \caption{Comparison between Aya, WildChat and~\datasetname~language distribution.}
  \label{fig:distribution}
\end{figure*}

\paragraph{Low-resource languages.}  Table~\ref{tab:low_resource} shows the results on $6$ languages. \datasetname~performs better than all the baselines across both Mistral-7B and LLama-3-8B. Specifically,~\datasetname~improves the performance by $1.3$ and $0.2$ on average for both models respectively. We also evaluate cross-lingual machine translation performance on extremely low resource languages as shown in Table~\ref{tab:flores200_eval}. Table~\ref{tab:flores200_eval} demonstrates that~\datasetname~outperforms existing baseline datasets by noticeable margin of $0.2$ BLeU score improvements with Mistral-7B and $0.5$ with LLama-3-8B. The M2Lingual models outperform all baselines in translating between different low-resource languages, except on Somhali, Awadhi (LLama-3-8B) and Yoruba (Mistral-7B) where their performance is a close second. This highlights a better coverage of low-resourced languages in~\datasetname~(Figure~\ref{fig:distribution}). 

\paragraph{Effect of IFT datasets on different sized LLMs.} We also study the impact of \datasetname~on a smaller scale model (QWEN-1.8B) and a larger model (LLaMA-2-13B). As shown in~\Cref{tab:different_sizedLLM_eval}, QWEN-1.8B on an average,~\datasetname~leads to $1.96$, $10$ and $4.5$ points improvements across MT-bench, QA and MGSM respectively. Similarly for the LLaMA-2-13B we get $0.8$, $3.75$ and $3.0$ points increase. The MT-Bench results across different languages are shown in~\Cref{tab:different_sizedLLM_eval_full} in Appendix.

\begin{figure*}[htbp!]
  \centering
  \includegraphics[width=\linewidth, height=.25\linewidth]{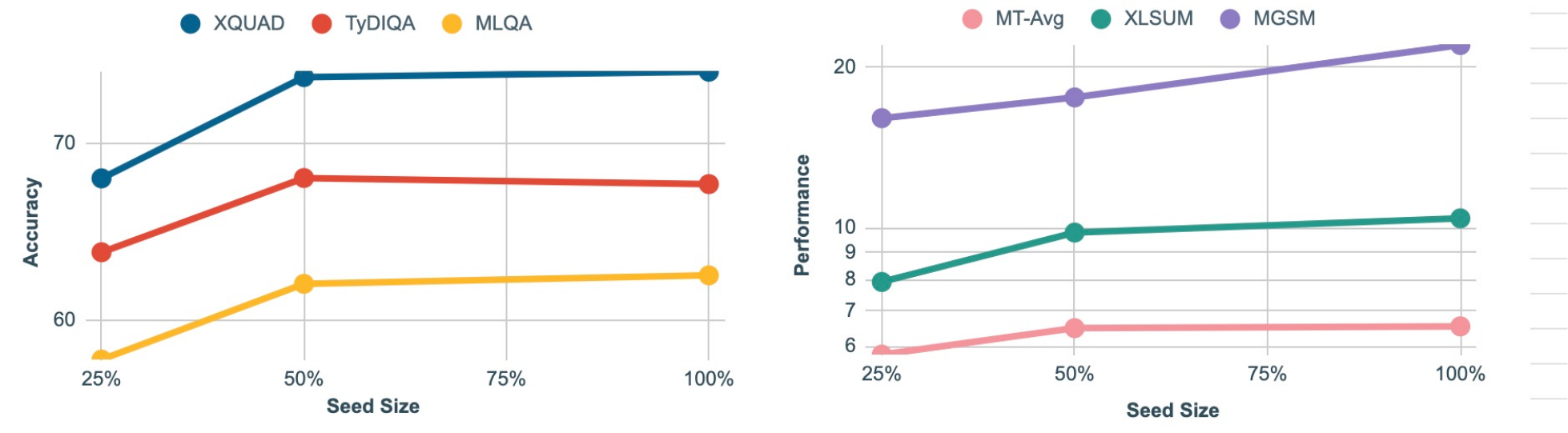}
  \caption{Performance vs seed size in data synthesis}
  \label{fig:combined_seed_performance_transposed}
\end{figure*}

\paragraph{Scaling data synthesis with seed size.}
Finally, to show how performance changes as we scale synthetic data generation on more seed examples only, we ran $2$ ablations where we 1) use only 25\% of seed examples and use its synthesized data and 2) use 50\% of the seeds. \Cref{fig:combined_seed_performance_transposed} shows that as we scale data synthesis by selecting more seeds the performance increases across all benchmarks. Specifically, on an average we see $0.5$ improvement in multilingual MT-bench, $5\%$ in QA and MGSM and $2.50$ in summarization. Additional analysis and tables for \Cref{fig:combined_seed_performance_transposed} are shown in \Cref{sec:without_seed_perf}.

\paragraph{Distribution of Languages.} \Cref{fig:distribution} shows a balanced representation of languages in our dataset compared to WildChat and Aya, which have uneven or very skewed distribution. This highlights a broader coverage of mid to low resource languages and explains the consistent performance improvements across high-mid resource languages in \Cref{tab:combined_evaluation,tab:combined_evaluation_data_in_wild,tab:mt_bench_eval,tab:evaluation} and low resource languages in Table~\ref{tab:low_resource} and~\ref{tab:flores200_eval}.

\paragraph{Token Lengths per Utterance.} \label{label:tok}
\Cref{tab:token_stats} shows that \datasetname~has one of the highest user and assistant turn tokens and the highest total number of tokens (computed via LLama tokenizer). This explains on complex benchmarks such as MT-Bench and MGSM, which require reasoning using a chain-of-thought or processing long contexts; and an occasional slight drop in performance on QA or summarization tasks, as the F1-score does not fully capture long, detailed answers.

\begin{table}[htbp!]
\centering
\small
\begin{tabular}{lccc}
\toprule
\textbf{Dataset} & \textbf{\#User tokens} & \textbf{\#Assistant} & \textbf{\# Total} \\ \midrule
Aya             & 157.9  & 447.2  & 605.1  \\
Bactrian X      & 114.34 & 354.07 & 468.41 \\
Multialpaca     & 47.58  & 97.22  & 144.8  \\
Open Assistant  & 37.54  & 277.22 & 314.7  \\
ShareGPT        & 99.97  & 558.06 & 658.03 \\
WildChat        & \textbf{282.33} & 442.25 & 724.58 \\
M2Lingual       & 200.28 & \textbf{558.86} & \textbf{759.14} \\ \bottomrule
\end{tabular}
\vspace{2mm}
\caption{Token statistics for different datasets}
\label{tab:token_stats}
\end{table}

\paragraph{Content moderation.} To ensure low toxicity in \datasetname's content as well as evaluate the \Evol~synthesis method's sensitivity in data generation, we conduct moderation testing with OpenAI Moderation API \cite{openaimoderation} following~\cite{zhao2024WildChat}. \Cref{tab:moderation-analysis} shows that less than $0.2\%$ of~\datasetname~is flagged by the Moderation API. We remove the flagged utterances before making the dataset public. It is worth noting that Human-AI generated datasets like WildChat have substantially more sensitive content.
\begin{table}[htbp!]
\centering
\small
\begin{tabular}{lccccccc}
\toprule
\textbf{Dataset} & \textbf{User\%} & \textbf{Chatbot\%} & \textbf{Avg\%} \\ 
\midrule
Alpaca & 0.01 & 0.02 & 0.01 \\ 
Aya & 0.10 & 0.24 & 0.22 \\ 
Open Assistant & 0.53 & 0.45 & 0.49 \\ 
Share GPT & 0.16 & 0.28 & 0.22 \\ 
WildChat & 6.05 & 5.18 & 5.61 \\ 
\datasetname & 0.21 & 0.13 & 0.17 \\ 
\bottomrule
\end{tabular}
\vspace{2mm}
\caption{Content moderation analysis reported from respective dataset papers (BactrainX does not perform toxicity analysis)}
\label{tab:moderation-analysis}
\end{table}
\section{Conclusion}
We introduce~\datasetname~- the \emph{first fully synthetic, multi-turn multilingual dataset} - containing $175K$ complex conversations across $70+$ languages and $19$ NLP tasks. We propose a scalable, cost-efficient and fully synthetic method for creating conversations using a two-step enrichment process based on the~\Evol~prompt taxonomy, which can be adapted to any task or monolingual data. Exhaustive experiments across \emph{three} model families and \emph{five} model sizes with evaluations spanning $31$ languages demonstrate the advantages of~\datasetname~over other datasets. Furthermore, our ablations and analysis on low-resource language support, content moderation, conversation length and language distribution demonstrate the quality of~\datasetname~over other datasets.

\section{Limitations and Ethical Considerations}

\datasetname~covers over 70 languages in total, with dialects added in with \Evol~as well - a significant number of languages, more than all relevant datasets (\Cref{tab:all-datasets}). However, there are many more languages in the real-world, and we cannot cover them all. We hope that our contribution helps expand access to languages, and future work can further build better access for all. Moreover, the performance of LLMs improves on low resource data with finetuning on \datasetname, showcasing the importance of including multiple languages and turns.

Some major limitations of \datasetname~include the limited conversation length, possible presence of toxic data, and dependence on GPT-4 translated MT-Bench for low-resource language evaluation. While potentially longer conversations could be built with \Evol, it would take significantly more resources to extend each conversation beyond the current limit. For toxicity, our seed dataset Aya does not contain specific flags for toxic, harmful, or offensive speech \cite{ayadataset}, and Aya authors report that they believe there is a low risk for these in Aya data. However, to mitigate risk, we conduct moderation analysis of the generated~\Evol~IR pairs for \datasetname, and find that less than $0.2\%$ of the generated data was flagged, which we filter out before making the data public. Lastly, we conduct limited manual evaluation of the GPT-4 generated low-resource multilignual MT-Bench data generated by GPT4, and find that it performs satisfactorily well. However, improving evaluation on low-resource data remains an area of future work.



\section*{References}
\bibliographystyle{ieeetran}
\bibliography{custom}

\section*{Appendix}
\newpage

\section{Appendix}
\subsection{Experiment Details}
We conduct experiments across \emph{three} model families \& \emph{five} model sizes --- Mistral-7B~\cite{jiang2023mistral}, LLaMA-3-8B~\cite{llamapaper} and QWEN-4B~\cite{bai2023qwen}. 
Furthermore, to demonstrate the effectiveness of our dataset across different model scales, we fine-tune both a larger model, LLaMA-2-13B~\cite{touvron2023llama}, and a smaller model, QWEN-1.8B~\cite{bai2023qwen}. To evaluate how well the datasets work with instruction-tuned models, we also experiment with Mistral-Instruct-7B.

\subsection{Baseline Datasets}
We use \emph{six} different multilingual datasets as baselines for comparison: 1) the top ranked conversation trees from \textbf{Open Assistant}~\cite{kopf2024openassistant},
2) \textbf{Aya}~\cite{ayadataset},
3) self-instruct dataset \textbf{MultiAlpaca}~\cite{wei2023polylm},
4) machine translated \textbf{Bactrian-X}~\cite{li2023bactrian} derived from Alpaca-52k~\cite{alpaca} and Dolly-15k~\cite{DatabricksBlog2023DollyV2},
5) the \textbf{ShareGPT}~\footnote{https://sharegpt.com/} collection, and
6) \textbf{WildChat}~\cite{zhao2024WildChat}.


For a fair comparison with WildChat, we use $200$K non-English conversations, ensuring the same language proportions, and downsampled $60$K English conversations, resulting in a total of $260$K conversations. Similarly for Bactrian-X, we sample $1$M IR pairs ensuring the same language proportions as in the original dataset.

\textbf{Additional Baselines} To highlight the importance of each step in our data curation process, we consider several ablations as baselines. Specifically we conduct experiments by training models using 1) only \textbf{Seed} samples, 2) seed samples with the generated evols (\textbf{Seed + Evol}) and 3) seeds, evols and the generated multi-turn conversations (\textbf{Seed + Evol + MT}). Finally, to see whether adding parallel data (PD) helps in improving the over model's performance, we collect $60$K from the Aya collection and train a baseline by augmenting the PD with our full dataset (\textbf{Seed + Evol + MT + PD}).
\subsection{Training}
All training is performed on $8$ A-$100$ $80$GB NViDIA GPUs \cite{choquette2021nvidia}, with the Axolotl\footnote{https://github.com/OpenAccess-AI-Collective/axolotl} framework. We used Mistral tags~\cite{jiang2023mistral} for finetuning all models. We use a batch size of $64$, a maximum sequence length of $8192$, a learning rate of $5 \times 10^{-6}$, the Adam optimizer \cite{kingma2014adam} with a cosine scheduler, and $10$ warmup steps. We reserve a $5$\% validation split, and train all the models until validation loss convergence. We compute the loss only on the targets using fp$16$ training.

\subsection{Evaluation}
\textbf{Multilingual benchmarks.} We utilize the EleutherAI evaluation framework \cite{eval-harness} for consistent comparisons. We evaluate the performance of different multilingual datasets on the following tasks:
  \begin{itemize}[noitemsep, nolistsep, left=0pt]
    \item \textit{Question Answering (QA)}: We focus on $3$ multilingual QA datasets 1) XQUAD~\cite{artetxe2019cross} with QA across $11$ languages, 2) TyDiQA~\cite{clark2020tydi} which has human generated QA in $11$ languages and 3) MLQA~\cite{lewis2019mlqa} with QA in $7$ languages. While QA data requires short answer phrases, conversational IR pairs might lead to longer answer span generation. Hence, we use $3$ in-context examples to get the right output format for LLMs. In the interest of time, we keep the number of examples per language to $100$ for XQUAD and MLQA, and $1000$ for TyDiQA. We use the validation set for XQUAD and test set for TyDiQA \& MLQA, and compute the standard F1-score.

    \item \textit{Summarization}: We use the XLSUM~\cite{hasan2021xl} dataset and focus on $6$ languages - Arabic, English, Spanish, French, Japanese and Russian. We restrict the total number of examples to $100$ and prompt the model to generate a summary in the same language as the context. We look at the ROUGE\textsubscript{L}~\cite{lin-2004-rouge} \& BLEU~\cite{papineni-etal-2002-bleu} scores for comparison.
    \item \textit{Classification}: We focus on XNLI~\cite{conneau2018xnli} and XCOPA~\cite{ponti2020xcopa} with $15$ and $11$ languages respectively in a zero-shot setting. We compute the accuracy (Acc) by looking at the log-likelihood assigned to the ground truth answer on the validation set.
    \item \textit{Multilingual math word problems}: We use MGSM~\cite{shi2022language}, a grade-school math benchmark that translates GSM8K~\cite{cobbe2021training} to $10$ different languages. Similar to QA tasks, we use $3$ in-context examples and compute the exact match (EM) with the ground truth answer.
  \end{itemize} 

\textbf{Translated MT-Bench.} To evaluate the conversation and instruction following ability of multilingual models across a wide array of tasks and languages, we translate MT-Bench~\cite{zheng2023judging}. MT-Bench comprises of $80$ multi-turn questions across $8$ domains. The models are required to respond to an initial and a follow-up question and GPT-4 assesses the model's responses on a scale of $1$ to $10$ ($10$ being the best), with the overall score being the mean over the two turns. We translate it into $9$ different languages with professional linguists to ensure high quality evaluation. We modify the judge prompt to include the language of the question asked at each turn, and additionally instruct GPT-4 to make sure the responses are in the same language as the question asked. We report the average scores across all $80$ examples for each language and also report the average MT-Bench score across all languages.
\begin{table}[h!]
\centering
\small
\begin{tabular}{@{}ccccc@{}}
\toprule
\textbf{Model} 
& Benchmark  & \textbf{Aya-seeds} & \textbf{Seed + Evol} \\ 
\midrule 

\multirow{7}{*}{\rotatebox[origin=c]{90}{\textbf{Mistral-7B}}}  & MT-Avg     & 4.40         & \textbf{5.57}  \\
\cmidrule[0.1pt](lr){2-4}
   & XQUAD      & 70.40        & \textbf{71.01} \\
   & MLQA     & 56.10        & \textbf{57.47} \\
   & MGSM       & 15.32        & \textbf{18.38} \\
  & XNLI       & 40.77        & \textbf{43.00} \\
   & XCOPA      & 55.55        & \textbf{57.55} \\
\bottomrule
\end{tabular}
\vspace{2mm}
\caption{\datasetname~vs same size Aya-seeds (100K).}
\label{tab:more_ayaseed_comparison}
\end{table}

\begin{table}[h!]
\centering
\small
\begin{tabular}{lccc}
\toprule
 Benchmark& \textbf{No seeds} & \textbf{25\% seeds} & \textbf{100\% seeds} \\ 
 & 157K    & 160K    & 175K   \\
\midrule
MT-EN                 & 7.00     & 6.86     & 7.13   \\
MT-FR    & 6.87 (6.80)  & 6.75 (6.79)  & 6.75 (6.81) \\
MT-IT                 & 6.84     & 6.78     & 6.90   \\
MT-JP                 & 5.81     & 5.80     & 5.70   \\
MT-ES                 & 6.52     & 6.70     & 6.81   \\
MT-DE                 & 6.45     & 6.37     & 6.39   \\
MT-NL                 & 6.46     & 6.23     & 6.34   \\
MT-Avg                & 6.57     & 6.55     & 6.54   \\
\midrule
XQUAD                 & 71.49    & 72.96    & 74.53  \\
TyDIQA                & 68.09    & 69.09    & 67.57  \\
MLQA                  & 58.57    & 61.62    & 62.40  \\
XLSUM        & 9.38     & 9.45     & 10.42  \\
MGSM                  & 17.95    & 18.25    & 22.0  \\
\bottomrule
\end{tabular}
\vspace{2mm}
\caption{Mistral-7B results with variable seeds on all benchmarks.}
\label{tab:combined_seed_performance_transposed}
\end{table}
\begin{table}[h!]
\small
\centering
\begin{tabular}{@{}lccc@{}}
\toprule
\textbf{Dataset} & \textbf{25\%} & \textbf{50\%} & \textbf{100\%} \\ \midrule
\textbf{MT-Avg}  & 5.79          & 6.49          & 6.54           \\
\midrule
\textbf{XQUAD}   & 67.91         & 74.18         & 74.53          \\
\textbf{TyDIQA}  & 63.67         & 67.93         & 67.57          \\
\textbf{MLQA}    & 57.98         & 61.94         & 62.40          \\
\textbf{XLSUM}   & 7.92          & 9.80          & 10.42          \\
\textbf{MGSM}    & 16.05         & 17.55         & 22.00          \\ \bottomrule
\end{tabular}
\vspace{2mm}
\caption{Mistral-7B performance results across benchmarks with different seed sizes used in figure~\ref{fig:combined_seed_performance_transposed}.}
\label{tab:var_seed}
\end{table}

\subsection{Importance of synthetic~\Evol s}
\label{sec:evol_seed}
To assess whether the importance lies in the use of synthetic~\Evol s rather than simply increasing the amount of seed-like data, we sampled an additional $94.9K$ IR pairs from the Aya dataset and collection and replace it with synthetic generated~\Evol s. Results in~\ref{tab:more_ayaseed_comparison} show that without synthetic~\Evol s the performance decreases, whereas having the same number of~\Evol~IR pairs leads to higher performance especially in MT-Bench and MGSM by $1.1$ and $3.38$ points respectively.

\subsection{~\datasetname~performance without seed examples}
\label{sec:without_seed_perf}
Table~\ref{tab:combined_seed_performance_transposed} demonstrates the performance of our dataset $(1)$ without seeds and $(2)$ with 25\% seed examples. Results show strong performance with multilingual MT bench without any seeds. It improves slightly compared to the last column that has all seed examples. The performance on other benchmark drops slightly but it still outperforms the evaluated baseline datasets in the paper.
\subsection{Results with variable seed size}
Finally, to show how performance changes as we scale synthetic data generation on more seed examples only, we ran $2$ ablations where we $(1)$ use only 25\% of seed examples and use its synthesized data and $(2)$ use 50\% of the seeds. Figure~\ref{fig:combined_seed_performance_transposed} and Table~\ref{tab:var_seed} demonstrate that as we scale data synthesize by selecting more seed examples the performance increases across all benchmarks. Specifically, on an average we see $0.5$ improvement in multilingual MT-bench, $5\%$ in QA and MGSM and $2.50$ in summarization.

\subsection{Complete Results }
\label{sec:qweb4b}
Tables~\ref{tab:mt_bench_eval},~\ref{tab:evaluation},~\ref{tab:evaluation_instruct} and~\ref{tab:mt_bench_eval_instruct} shows the complete results comparing~\datasetname~against all the baseline datasets, $4$ base models across multilingual MT-Bench, question answering, summarization and classification tasks. Table~\ref{tab:different_sizedLLM_eval_full} compares~\datasetname~against top performing baseline on a smaller (Qwen1.8B) and a larger (LLama-2-13B) model. 
\paragraph{QWEN-4B \& Mistral-Instruct-7B results}
We evaluate Mistral-Instruct-7B to highlight the impact of multilingual IFT datasets on pre-instruction finetuned models. \datasetname~leads Mistral-Instruct-7B to achieve best performance in $5$ of $8$ MT-Bench language evaluations and $5$ of the $7$ multilingual evaluation benchmarks as shown in Tables \ref{tab:evaluation_instruct} and \ref{tab:mt_bench_eval_instruct} respectively. Interestingly, the improvements from \datasetname~in Mistral-Instruct-7B over baseline datasets is consistently higher when compared to Mistral-7B-base (\Cref{tab:evaluation}) in all of the multilingual QA tasks, MGSM, and XCOPA. 
We also evaluate QWEN-4B model to showcase results from smaller LLM from different model family. We observe similar findings as QWEN-4B finetuned with \datasetname~achieves competitive results in both MT-Bench and multilingual evaluation datasets. Another interesting observation is that improvements seem relatively higher for QWEN-4B model using \datasetname~when compared to Mistral-7B and LLaMA-3-8B models, highlighting the usefulness of our proposed data on moderate sized LLMs. 

\definecolor{darkgreen}{HTML}{afc8b5}
\definecolor{lightgreen}{HTML}{c5eace}

\begin{table*}[htbp!]
  \centering
  \small
  \begin{adjustbox}{max width=\textwidth}
  \begin{tabular}{@{}c@{}cccccccccc@{}}
    \toprule
    \textbf{Model} & \textbf{Dataset}  & \textbf{MT-EN} & \textbf{MT-FR} & \textbf{MT-IT} & \textbf{MT-JP} & \textbf{MT-ES} & \textbf{MT-DE} & \textbf{MT-NL} & \textbf{MT-PT} & \textbf{MT-AVG} \\
    \midrule
    \multirow{9}{*}{\rotatebox[origin=c]{90}{\textbf{Mistral-7B}}}
    & Open Assistant & 6.72   & 5.87 (5.90)   & 6.04   & 4.19   & 5.87   & 5.82   & 4.97 & 6.01 & 5.66  \\
    & MultiAlpaca  & 5.45   & 4.90 (5.22)   & 4.63   & 3.76   & 5.01   & 4.66   & 4.51 & 4.65 & 4.77   \\
    & Bactrian-X   & 5.60   & 5.35 (5.26)   & 5.46   & 4.82   & 5.24   & 5.53   & 4.96 & 5.31 & 5.25  \\
    & ShareGPT    &  7.04   & 5.93 (5.70)   & 5.42   & 4.75   & 5.83   & 6.00   & 5.27  & 5.92 & 5.80  \\
    & WildChat    & 7.02   &  6.46 (6.77)   &  6.68   &  5.50  &  6.71  & \textbf{6.43}   & \textbf{6.51}  & \textbf{6.89} &  6.53  \\
    & Aya    & 6.43   & 5.42 (5.39)   & 4.97  & 3.37   & 5.45   & 5.37   & 4.94  & 5.12 & 5.18  \\
     \cmidrule[0.1pt](lr){2-11}
    & Seed      & 6.01   & 5.15 (5.14)   & 5.35   &  3.44  & 5.07   & 5.98   & 4.62 & 4.91 & 5.04  \\
    & Seed + Evol & 6.33   & 5.44 (5.30)   & 5.46   & 4.74  & 5.88   & 5.61   & 5.40  & 5.78 & 5.56  \\
    & Seed + Evol + MT (\datasetname) & \textbf{7.13} & \textbf{6.75 (6.81)}   & \textbf{6.9}   & \textbf{5.70}   & \textbf{6.81}   &  6.39  &  6.34 &  6.46 & \textbf{6.54}  \\
    \midrule
    \multirow{9}{*}{\rotatebox[origin=c]{90}{\textbf{LLaMA-3-8B}}}
    & Open Assistant & 6.26  &  5.15 (5.03)  & 4.95  & 4.08 & 5.26   & 4.87   & 5.01 & 5.48 & 5.12 \\
    & MultiAlpaca  & 4.96   & 4.60 (5.09)   & 4.22   & 3.30   & 4.76   & 4.18   & 4.32  & 4.27  & 4.41 \\
    & Bactrian-X   & 6.27   & 5.73 (5.77)   & 5.73   & 4.83   & 5.95   & 5.34   & 5.41  & 5.90 & 5.66 \\
    & ShareGPT    & 7.07   & 6.17 (5.76)   & 6.43   & 5.40   & 6.10   & 6.07  & 5.82 & 6.13  &  6.10 \\
    & WildChat    &  \textbf{7.20}   &  \textbf{6.74 (6.96)}   &  6.78   &  \textbf{6.35}   &  6.86   &  6.60   &  6.58  &  6.72 &  \textbf{6.75}  \\
    & Aya    & 5.95   & 5.01 (4.50)   & 5.41   & 3.86   & 5.27   & 4.93  & 4.66 & 4.95 & 4.95  \\
    \cmidrule[0.1pt](lr){2-11}
    & Seed      &  4.38  & 3.55 (3.75)   & 3.56   & 2.68   & 3.52   & 3.42   & 3.45 & 3.54 & 3.54  \\
    & Seed + Evol & 6.95   & 6.41 (6.50)   & 6.22   & 5.41   & 6.35   & 6.11   & 5.90  & 5.27 & 6.12  \\
    & Seed + Evol + MT (\datasetname) &  7.17 & 6.55 (6.82)   &  \textbf{6.86}   & 6.26   &  \textbf{6.95}   &  \textbf{6.65}   &  \textbf{6.93}  &  \textbf{6.81} &  6.74  \\
    \bottomrule
  \end{tabular}
  \end{adjustbox}
  \caption{Multilingual MT-Bench results. Canadian French results are in MT-FR brackets. Best scores are in bold and dark green while 2\textsuperscript{nd} best are in light green. \textit{Seeds} are $15.1$K seeds; \textit{Seed + Evol} is additional \Evol~IR pairs. \textit{Seed + Evol + MT} has additional multi-turn data.}
  \label{tab:mt_bench_eval}
\end{table*}

\definecolor{darkgreen}{HTML}{afc8b5}
\definecolor{lightgreen}{HTML}{c5eace}

\begin{table*}[h!]
  \centering
  \small
  \begin{adjustbox}{max width=\textwidth}
  \begin{tabular}{@{}c@{}cccccccccc@{}}
    \toprule
    
    \multirow{2}{*}{\textbf{Model}} & \multirow{2}{*}{\textbf{Dataset}}  & \textbf{XQUAD} & \textbf{TyDiQA} & \textbf{MLQA} & \multicolumn{2}{c}{\textbf{XLSUM}} & \textbf{MGSM} & \textbf{XNLI} & \textbf{XCOPA} \\
     &  & F1 & F1 & F1 & ROUGE\textsubscript{L} & BLEU & EM & Acc & Acc \\
    \midrule
    \multirow{9}{*}{\rotatebox[origin=c]{90}{\textbf{Mistral-7B}}}
    & Open Assistant & 67.99   & 54.22  & 53.64   & 10.86 & 0.85   & 16.05   & 42.74   & 56.73   \\
    & MultiAlpaca  & 67.99   & 64.44   & 55.69   & 10.9 & 1.59   & 10.41   & 42.18   & 58.91  \\
    & Bactrian-X   & 71.91   & 66.63   &  60.27   & 3.30 & 0.20   &  17.14   &  \textbf{43.91}   & 58.64   \\
    & ShareGPT  & 66.33   & 56.97   & 50.78   & 3.31& 0.288   & 11.32   & 41.13   & 56.09   \\  
    & WildChat    &  72.55   & 64.27   & 59.53   & 3.91 & 0.41   &  {18.41}   & 43.11  & 58.00   \\
    & Aya    & 70.46   &  66.95   & 57.47   &  \textbf {12.5} &  \textbf{2.01}   & 13.86   & 41.78  & 59.00  \\
    \cmidrule[0.1pt](lr){2-10}
    & Seed   & 72.52  & 65.89  & 59.33 &  11.53 &  1.72  &  13.65  &  42.28  & 57.64  \\
    & Seed + Evol & 71.01   & 65.04   & 57.47   & 9.8 & 1.37  & 18.38  & 43.00   & 57.55 \\
    & Seed + Evol + MT (\datasetname) &  \textbf{74.53} &  \textbf{67.57}  &  \textbf {62.40}  & 10.42 & 1.38  & \textbf{22.00}  & 42.12   &  \textbf{59.55}   \\
    \midrule
    \multirow{9}{*}{\rotatebox[origin=c]{90}{\textbf{LLaMA-3-8B}}}
    & Open Assistant & 64.38   & 52.65  & 47.08  & 9.38 & 1.21   & 17.36   & 46.17   &  \textbf{63.82}   \\
    & MultiAlpaca  & 75.08   & 64.49   & 59.01  &  \textbf{10.98} &  \textbf{1.45}   & 10.68   &  \textbf{46.93}   &  63.55   \\
    & Bactrian-X   & 69.57  &  56.45 & 58.51  & 8.39 & 1.28   & 22.86   &  46.90   & 62.18   \\
    & ShareGPT    & 56.98   & 58.48   & 43.43 & 3.53 & 0.40  & 25.32   & 45.93   & 63.00   \\
    & WildChat    & 63.15   & 59.88  &  63.16   & 5.52 & 0.76   & 26.36   & 46.88   & 62.27   \\
    & Aya    & 75.14  & 59.60   & 53.14   &  10.38 &  1.39   & 22.09   & 45.64   &  63.55   \\
    \cmidrule[0.1pt](lr){2-10}
    & Seed      &  \textbf{77.27}   &  68.57  & 60.01  & 9.92 &  \textbf{1.45}   & 17.18   & 46.02   & 62.82   \\
    & Seed + Evol & 76.17  &  \textbf{69.89}   & 63.09   &   8.96 & 1.23 &  \textbf{28.00}   & 46.38   & 61.36   \\
    & Seed + Evol + MT (\datasetname) & 75.91 & 67.84  &  \textbf{63.50}   & 8.87 & 1.25   & 27.36   & 46.18   & 62.55   \\
    \bottomrule
  \end{tabular}
  \end{adjustbox}
  \caption{Evaluations of LLaMA-3-8B-base \& Mistral-7B-base in different tasks. Same notations as in Table \ref{tab:mt_bench_eval}}
  \label{tab:evaluation}
\end{table*}

\begin{table*}[h!]
  \centering
  \begin{adjustbox}{max width=\textwidth}
  \begin{tabular}{@{}c@{}ccccccccccc@{}}
    \toprule
    \multirow{2}{*}{\textbf{Model}} & \multirow{2}{*}{\textbf{Dataset}}  & \textbf{XQUAD} & \textbf{TyDiQA} & \textbf{MLQA} & \multicolumn{2}{c}{\textbf{XLSUM}} & \textbf{MGSM} & \textbf{XNLI} & \textbf{XCOPA} & \multirow{2}{*}{\textbf{MT-Avg}} \\
    &  & F1 & F1 & F1 & ROUGE\textsubscript{L} & BLEU & EM & Acc & Acc \\
    \midrule
    \multirow{11}{*}{\rotatebox[origin=c]{90}{\textbf{QWEN-4B}}}
    & Open Assistant & 53.63   & 45.30   & 46.34   & 4.15 & 0.29   & 17.50   &  38.52   &  58.45 & 3.47 \\
     & MultiAlpaca  & 51.81  &  53.51  & 40.26   & 8.9 & 1.0   & 12.1   & 38.3   & 58.40 & 2.93 \\
    & Bactrian-X   & 46.70   & 42.79   & 42.2   & 7.1 & 0.8   & 18.6   & 38.3   & 57.70 & 3.80  \\
    & ShareGPT    & 41.86   & 28.20  & 36.03   & 4.58 & 0.43   & 16.95   & 37.83   &  \textbf{58.55}  & 3.80 \\
    & WildChat    & 53.18   & 49.18  & 42.81   & 5.23 & 0.56   & 19.27   &  \textbf{38.74}   & 58.18  &  \textbf{4.29} \\
    & Aya    & 54.00   & 52.14   & 48.28   &  \textbf{10.91} &  \textbf{1.31}   & 16.50   & 37.59   & 57.73 & 3.43  \\
    \cmidrule[0.1pt](lr){2-11}
    & Seed      &  \textbf{66.55}  &  \textbf{58.09}   & 48.25  &  10.65 & 0.65   & 15.36   & 37.59   & 58.00  & 2.47 \\
    & Seed + Evol & 52.24$^{\star}$   & 52.50  &  49.87   & 8.50 & 1.12   &  20.77   & 38.36   & 57.91  & 3.79 \\
    & Seed + Evol + MT (\datasetname) & 49.12$^{\star}$  & 47.53  &  \textbf{50.36}  & 8.30 & 1.02  &  \textbf{21.36}     & 38.37   & 58.36 & 4.23  \\
    \midrule
    \multirow{11}{*}{\rotatebox[origin=c]{90}{\textbf{Mistral-Instruct-7B}}}
    & Open Assistant & 61.33   & 59.28   & 53.27  & 9.62 & 1.43   & 19.00   &  43.91   & 58.09 & 5.58  \\
    & MultiAlpaca  & 63.76  & 63.05   & 51.09   & 11.51 & 1.80   & 13.18   &  \textbf{44.70}   & 58.18  & 4.74 \\
    & Bactrian-X   & 70.5   & 64.8   & 50.60   & 9.14 & 1.35   & 17.91   & 42.23   & 57.25  & 5.98 \\
    & ShareGPT    & 44.53   & 49.5   & 40.45   & 3.31 & 0.38   & 17.36   & 42.13   & 56.73 & 6.11  \\
    & WildChat    & 61.53   & 53.1   & 52.60   & 6.31 & 0.56   & 21.00   & 41.86   & 57.75 &  6.62  \\
    & Aya    & 69.9   & 66.43  & 57.27   &  \textbf{12.58} &  \textbf{2.05}   & 16.36   & 42.84   &  58.60  & 5.20 \\
    \cmidrule[0.1pt](lr){2-11}
    & Seed      & 68.78   & 61.54   & 56.11   & 12.45 &  2.04   & 18.27   & 43.23   & 58.45  & 3.92 \\
    & Seed + Evol &  \textbf{72.87}   & 68.43   & 55.43   &  12.51 & 1.33   &  22.00   & 42.51   & 58.09 & 6.48  \\
    & Seed + Evol + MT (\datasetname) &  71.41 &  69.44   &  58.33   & 9.57 & 1.51   & 19.82   & 42.37   &  \textbf{59.45} &  \textbf{6.64}  \\
    \bottomrule
  \end{tabular}
  \end{adjustbox}
   \vspace{0.15cm}
  \caption{Evaluations of QWEN-4B \& Mistral-Instruct-7B in different tasks and MT-Bench score averaged across languages. Please see \cref{tab:mt_bench_eval_instruct} in appendix for MT-Bench score in each language. $\star$ in XQUAD, TyDiQA scores for QWEN-4B show exception cases where outputs had repeated noisy patterns in multiple runs resulting in low scores.}
  \label{tab:evaluation_instruct}
\end{table*}

\begin{table*}[htbp!]
  \centering
  \begin{adjustbox}{max width=\textwidth}
  \begin{tabular}{@{}c@{}cccccccccc@{}}
    \toprule
    \textbf{Model} & \textbf{Dataset}  & \textbf{MT-EN} & \textbf{MT-FR} & \textbf{MT-IT} & \textbf{MT-JP} & \textbf{MT-ES} & \textbf{MT-DE} & \textbf{MT-NL} & \textbf{MT-PT} & \textbf{MT-Avg} \\
    \midrule
    \multirow{11}{*}{\rotatebox[origin=c]{90}{\textbf{QWEN-4B}}}
    & Open Assistant & 5.95   & 3.49 (3.66)   & 2.84   & 2.38   & 3.88   & 2.73   & 2.46 & 3.23  & 3.47 \\
    & MultiAlpaca  & 4.74   & 3.29 (2.88)  & 2.65   & 1.90   & 3.15   & 2.56   & 2.08  & 2.90  & 2.93 \\
    & Bactrian-X   & 5.88   & 3.84 (4.03)  & 3.25   & 2.66   & 3.85   & 3.49   & 2.77  & 3.90 & 3.80 \\
    & ShareGPT    & 5.89   & 3.92 (4.02)   & 3.39  & 3.13   & 4.20   & 2.97   & 2.55 & 3.72  & 3.80 \\
    & WildChat    & 6.27   & 4.49 (4.81)   & 3.83   & 3.20   & 4.38   & 3.83  & 3.11 & 4.27  & 4.29 \\
     & Aya    & 5.24   & 3.45 (3.74)   & 2.96   & 2.24   & 3.77  & 3.08   & 2.44  & 3.51 & 3.43 \\
    \cmidrule[0.1pt](lr){2-11}
    & Seed      & 4.60   & 2.68 (2.63)   & 2.09   & 1.59   & 2.43   & 2.18   & 1.67  & 2.03 & 2.47 \\
    & Seed + Evol & 5.81   & 3.86 (4.03)   & 3.00   & 2.82   & 4.24   & 3.35   & 2.53 & 3.68  & 3.79 \\
    & Seed + Evol + MT & 6.01 & 4.67 (4.62)   & 3.55   & 3.36   & 4.48   & 3.83  & 2.89 & 4.02  & 4.23 \\
    \midrule
    \multirow{11}{*}{\rotatebox[origin=c]{90}{\textbf{Mistral-Inst 7B}}}
    & Open Assistant & 6.76   & 5.74 (6.07)   & 5.73   & 3.78   & 5.84   & 5.91   & 4.99 & 5.60  & 5.58 \\
   & MultiAlpaca  & 5.90   & 4.83 (4.82)   & 4.66   & 3.25  & 5.01   & 4.57   & 4.84 & 4.71 & 4.74  \\
    & Bactrian-X   & 7.06  &  5.96 (6.02)  & 6.22 & 4.53 & 6.25  & 6.09  & 5.81 & 6.15 & 5.98 \\
    & ShareGPT    & 6.84   & 6.34 (6.20)  & 5.84   & 4.61   & 6.51   & 6.10   & 6.06  & 6.25  & 6.11 \\
    & WildChat  & 7.39 & 6.77 (6.53) & 6.737 & 5.64 & 6.503 & 6.80 & 6.39 & 6.95 & 6.62 \\
    & Aya    & 5.83   & 5.32 (5.78)   & 5.45   & 3.61   & 5.39   & 5.06   & 5.28 & 5.32  & 5.20 \\
    \cmidrule[0.1pt](lr){2-11}
    & Seed      & 4.85   & 4.28 (4.24)   & 3.98  & 2.44   & 3.98   & 3.71   & 3.85 & 4.03   & 3.92 \\
    & Seed + Evol & 7.20 & 6.24 (6.56) & 6.40 & 5.55 & 6.83 & 6.41 & 6.51 & 6.57 & 6.48\\
    & Seed + Evol + MT & 7.47 & 6.70 (6.50)  & 6.71 & 5.75 & 6.91 & 6.52 &  6.37  & 6.83 & 6.64 \\
    \bottomrule
  \end{tabular}
  \end{adjustbox}
  \caption{MT-Bench evaluations in different languages for QWEN-4B and Mistral-Instruct-7B.}
  \label{tab:mt_bench_eval_instruct}
\end{table*}

\newpage

\begin{table*}[h!]
\small
\centering
\resizebox{\textwidth}{!}{%
\begin{tabular}{@{}cccccccccccccc@{}}
\toprule
Model Name &   Dataset & MT-EN & MT-FR & MT-IT & MT-JP & MT-ES & MT-DE & MT-NL & MT-PT & MT-Avg & MGSM & MLQA & TyDiQA \\
\midrule
\multirow{3}{*}{Qwen-1.8B} & WildChat   &  4.99 &  2.75 (2.74) &  2.08 &  1.72 &  2.88 &  1.92 &  1.54 &  2.68 &  2.59 & 8.00 & 29.39 &  42.42 \\
              & MultiAlpaca &  3.97 &  1.93 (1.99) &  1.74 &  1.44 &  1.87 &  1.86 &  1.52 &  1.91 &  2.03 & 7.45 & 19.30 &  33.38 \\
              & \datasetname~&  \textbf{6.20} &  \textbf{4.55} (\textbf{4.25}) &  \textbf{3.85} &  \textbf{3.27} &  \textbf{4.40} &  \textbf{4.11} &  \textbf{3.32} &  \textbf{4.51} &  \textbf{4.27} &  \textbf{12.23} &  \textbf{38.24} &  \textbf{47.66} \\
\midrule
\multirow{3}{*}{LLaMA-2-13B} & WildChat   &  \textbf{6.64} &  6.25 (5.89) &  5.98 &  5.10 &  \textbf{6.20} &  \textbf{6.10} &  5.82 &  5.99 &  6.00 & 9.95 & 53.69 &  60.14 \\
              & MultiAlpaca &  5.09 &  4.35 (4.55) &  4.35 &  3.52 &  4.47 &  4.54 &  4.69 &  4.62 &  4.46 & 7.80 & 48.74 &  59.46 \\
              & \datasetname~&  6.47 &  \textbf{6.40} (\textbf{6.20}) &  \textbf{6.13} &  \textbf{5.35} &  6.18 &  5.94 &  \textbf{5.87} &  \textbf{6.17} &  \textbf{6.08} & \textbf{11.95} & \textbf{54.64} &  \textbf{64.66} \\
\bottomrule
\end{tabular}%
}
\caption{Evaluations of QWEN-1.8B and LLaMa-2-13B for highlighting impact on different sized LLMs.}
\label{tab:different_sizedLLM_eval_full}
\end{table*}
\subsection{Examples of Generated Evols and conversations}
\label{sec:examples}

\begin{table*}[h!]
\centering
\resizebox{\textwidth}{!}{
\begin{tabular}{@{}c p{9cm} p{3cm}@{}}
\toprule
\textbf{Step} & \textbf{Content} & \textbf{Evol} \\
\midrule
Seed Sample & What is the longest-running Broadway show ever? & \\
\midrule
evoled (Turn 1) Instr. & Which Broadway show holds the record for the longest run? Please answer briefly. & Concise \\
\midrule
GPT-4 Response & The Phantom of the Opera holds the record for the longest run on Broadway. & \\
\midrule
Turn 2 & Now, break down the following multi-part question and answer each part: Who wrote this show? What are some of its most popular songs? When did it first premiere on Broadway? & Expanding Queries \newline Change Context \\
\midrule
GPT-4 Response & 1. Who wrote this show?\newline
The Phantom of the Opera was composed by Andrew Lloyd Webber. The lyrics were written by Charles Hart and Richard Stilgoe.\newline\newline 
2. What are some of its most popular songs?\newline
Some of its most popular songs include "The Phantom of the Opera," "Music of the Night," "All I Ask of You," and "Masquerade."\newline\newline 
3. When did it first premiere on Broadway?\newline
The Phantom of the Opera first premiered on Broadway on January 26, 1988. & \\

\bottomrule
\end{tabular}}
\caption{Conversation example from \datasetname}
\end{table*}

\begin{table*}[h!]
\centering
\resizebox{\textwidth}{!}{
\begin{tabular}{@{}c p{9cm} p{3cm}@{}}
\toprule
\textbf{Step} & \textbf{Content} & \textbf{Evol} \\
\midrule
Seed Sample & Formulate an answer to this elaborate question: Where is the regional airline based that Jim French is the former chairman and CEO of? & \\
\midrule
evoled (Turn 1) Instr. & Formulate an answer to this elaborate question: Identify the base location of the regional airline for which Jim French served as the former chairman and CEO, and explain the significance of this location in the context of the airline's regional operations. & Combine Facts \\
\midrule
GPT-4 Response & Jim French served as the former chairman and CEO of Flybe, a British regional airline. The base location of Flybe was Exeter International Airport in Devon, England. \newline\newline The significance of Exeter International Airport as the base location for Flybe lies in its strategic positioning within the United Kingdom. Exeter is situated in the southwest region of England, which allowed Flybe to efficiently connect various regional destinations across the UK and Europe. This location provided a central hub for operations, facilitating shorter flight routes and quick turnaround times, which are essential for regional airlines aiming to offer frequent and reliable services. Additionally, having a base in Exeter helped Flybe tap into a market with fewer competing airlines, allowing it to establish a strong regional presence and foster economic growth within the southwest of England by improving connectivity and accessibility. & \\
\midrule
Turn 2 & Please adjust the output to provide the answer in a concise text-only format. & Adjust Output Format \\
\midrule
GPT-4 Response & Jim French served as the former chairman and CEO of Flybe, a British regional airline based at Exeter International Airport in Devon, England. The significance of this location lies in its strategic positioning in the southwest of England, enabling efficient regional connectivity across the UK and Europe and fostering economic growth in the region. & \\
\midrule
Turn 3 & Now, rephrase your response as if you were an enthusiastic travel blogger sharing fascinating details with your readers. & Persona Rewriting Follow Up \\
\midrule
GPT-4 Response & Hey travel enthusiasts! Did you know that Jim French, the former chairman and CEO of Flybe, once led this fantastic British regional airline based at Exeter International Airport in beautiful Devon, England? It's such a cool spot! Nestled in the southwest of England, Exeter’s prime location made it perfect for Flybe to connect countless regional destinations across the UK and Europe. This strategic hub wasn’t just about convenience; it played a crucial role in boosting regional connectivity and economic growth. So, next time you’re flying through Exeter, remember the incredible role it played in the world of regional air travel! & \\
\bottomrule
\end{tabular}}
\caption{Conversation from \datasetname}
\label{tab:formulate_answer_rephrase}
\end{table*}

\begin{table*}[h!]
\centering
\resizebox{\textwidth}{!}{
\begin{tabular}{@{}c p{9cm} p{3cm}@{}}
\toprule
\textbf{Step} & \textbf{Content} & \textbf{Evol} \\
\midrule
Seed Sample & Given the following statement, predict the tags for each word: Daniel Smith daniel.smith2@durham.ac.uk Durham University, United Kingdom & \\
\midrule
evoled (Turn 1) Instr. & Identify the appropriate tags for each element in this data: Daniel Smith daniel.smith2@durham.ac.uk Durham University, United Kingdom. Please provide a brief response. & Concise \\
\midrule
GPT-4 Response & Name: Daniel Smith\newline
Email: daniel.smith2@durham.ac.uk\newline
Institution: Durham University\newline
Country: United Kingdom & \\
\midrule
Turn 2 & Now, please reformat the output to display only the tags and associated elements in a structured JSON format. & Adjust Output Format \\
\midrule
GPT-4 Response & \texttt{\{ "Name": "Daniel Smith", "Email": "daniel.smith2@durham.ac.uk", "Institution": "Durham University", "Country": "United Kingdom" \}} & \\
\bottomrule
\end{tabular}}
\caption{Conversation from \datasetname}
\label{tab:identify_tags_conversation}
\end{table*}

\begin{table*}[h!]
\centering
\resizebox{\textwidth}{!}{
\begin{tabular}{@{}c p{9cm} p{3cm}@{}}
\toprule
\textbf{Step} & \textbf{Content} & \textbf{Evol} \\
\midrule
Seed Sample & Finish the following sentence with the best choice: To clean away the excess plastic from the edge of a project, you can\newline\newline Choices:\newline - Use a belt sander to remove the spare plastic\newline - Use a coffee grinder to remove the spare plastic\newline\newline Answer: & \\
\midrule
evoled (Turn 1) Instr. & Complete the sentence with the most appropriate choice: To prevent the documents from flying around in a windy environment, you can\newline\newline Choices:\newline - Use a heavy paperweight to keep them in place\newline - Use a feather to keep them in place\newline\newline Answer: & Object Properties \\
\midrule
GPT-4 Response & Use a heavy paperweight to keep them in place & \\
\midrule
Turn 2 & Now, generate a paragraph explaining the pros and cons of using a heavy paperweight to keep documents in place in windy conditions. & Challenging Follow Up\\
\midrule
GPT-4 Response & Using a heavy paperweight to secure documents in a windy environment has several advantages. First, the weight of the paperweight can effectively prevent papers from being blown away, ensuring they stay in the designated area. It also allows for easy accessibility and organization of the documents as they can be quickly retrieved without the hassle of chasing loose papers. In addition, paperweights are available in various designs and materials, providing aesthetic appeal to your workspace.\newline\newline However, there are also some potential drawbacks. The primary disadvantage is the risk of damage to the documents. If the paperweight is too heavy or it is dropped accidentally, it could tear or crumple the papers. Furthermore, paperweights take up desk space and can be inconvenient if you're working with a limited area. Lastly, if the wind is extremely strong, even a heavy paperweight might not be enough to keep the documents in place. & \\
\midrule
Turn 3 & Now, transform your previous response into a haiku format. The haiku should have a 5-7-5 syllable structure. Reconstruct your answer if necessary to fit this format. & Random Follow Up \\
\midrule
GPT-4 Response & Paperweight holds firm,\newline Yet may damage precious sheets,\newline In strong gusts, it fails. & \\
\bottomrule
\end{tabular}}
\caption{Conversation from \datasetname}
\label{tab:complete_sentence_haiku}
\end{table*}

\begin{table*}[h!]
\centering
\scriptsize
\resizebox{\textwidth}{!}{
\begin{tabular}{c p{9cm} p{3cm}}
\toprule
\textbf{Step} & \textbf{Content} & \textbf{Evol Type} \\
\midrule
Seed Sample & Quelle pourrait être la principale contribution de Dante, Boccacio et Petrach à la langue italienne? Étant donné la question précédente, écrivez un contexte qui contient la réponse. Cela peut être de 1 à 20 phrases. Le contexte: & \\
\midrule
Evolved (Turn 1) Instr. & Comment les œuvres de Dante, Boccace et Pétrarque ont-elles influencé non seulement la langue italienne, mais aussi la philosophie et la littérature européennes ? Étant donné cette question, écrivez un contexte qui contient la réponse. Cela peut être de 1 à 20 phrases. Le contexte: & Increase Reasoning \\
\midrule
GPT-4 Response & Les œuvres de Dante, Boccace et Pétrarque ont profondément marqué la langue italienne et ont eu une influence durable sur la philosophie et la littérature européennes. Dante Alighieri, avec sa "Divine Comédie", a non seulement consolidé l'usage du toscan comme langue littéraire, mais a également introduit des thèmes philosophiques et théologiques d'une grande profondeur, influençant la pensée médiévale et renaissante. Giovanni Boccace, à travers son "Décaméron", a enrichi la prose italienne et a offert un aperçu critique et souvent satirique de la société de son temps, ouvrant la voie à une littérature plus réaliste et humaniste. Francesco Pétrarque, avec ses sonnets et poésies en langue vernaculaire, a non seulement perfectionné la forme du sonnet, mais a aussi contribué à l'idéal de l'humanisme, valorisant l'étude des textes classiques et la quête de la connaissance personnelle. Ensemble, ces auteurs ont jeté les bases de la Renaissance italienne et ont influencé des écrivains et des penseurs à travers toute l'Europe, de Chaucer à Shakespeare, en passant par Montaigne et Cervantès, renforçant l'idée que la littérature peut être un moyen puissant d'exploration philosophique et d'expression culturelle. & \\
\midrule
Turn 2 & Maintenant, en vous basant sur le contexte fourni, pouvez-vous détailler comment les œuvres de ces trois auteurs ont spécifiquement influencé la Renaissance italienne ? & Context Retention Follow Up \\
\midrule
GPT-4 Response & Les œuvres de Dante, Boccace et Pétrarque ont joué un rôle central dans le développement de la Renaissance italienne en redéfinissant les normes littéraires, linguistiques et philosophiques de l'époque. Dante Alighieri, avec sa "Divine Comédie", a non seulement démontré que le toscan pouvait être une langue littéraire de premier ordre, mais a également exploré des thèmes universels comme la justice, la rédemption et l'amour divin, qui sont devenus des préoccupations centrales de la pensée renaissante. Son usage du toscan a contribué à l'unification linguistique de l'Italie, jetant les bases d'une langue nationale.\newline Giovanni Boccace, par le biais de son "Décaméron", a offert une vision plus terrestre et réaliste de la société, abordant des thèmes tels que l'amour, la fortune et l'intelligence humaine. Ses histoires, souvent humoristiques et critiques, ont encouragé une approche plus humaniste et empirique de la vie, en réaction aux dogmes rigides du Moyen Âge. Boccace a également influencé la prose narrative, rendant les histoires accessibles à un public plus large et diversifié. Francesco Pétrarque, quant à lui, est souvent considéré comme le père de l'humanisme. Ses œuvres, notamment son "Canzoniere", ont popularisé le sonnet et ont mis l'accent sur l'expression individuelle et la quête de la connaissance personnelle. Pétrarque a redécouvert et valorisé les textes classiques, inspirant une redécouverte des œuvres de l'Antiquité gréco-romaine, ce qui est une pierre angulaire de la Renaissance. Son admiration pour la culture classique a encouragé les intellectuels italiens à réévaluer leurs propres traditions et à chercher à les harmoniser avec les idéaux antiques. Ensemble, ces auteurs ont transformé la culture italienne et ont ouvert la voie à un mouvement intellectuel et artistique qui allait révolutionner l'Europe, en mettant l'accent sur la dignité de l'homme, l'importance de la raison, et la beauté de la langue et de la littérature. Leurs œuvres ont servi de modèles et de sources d'inspiration pour les artistes, écrivains et penseurs de la Renaissance, consolidant ainsi leur place comme piliers de ce mouvement culturel et intellectuel. & \\
\bottomrule
\end{tabular}}
\caption{Conversation from \datasetname~in Dutch}
\label{tab:dante_boccace_petrarque}
\end{table*}

\begin{table*}[h!]
\centering
\resizebox{\textwidth}{!}{
\begin{tabular}{c m{10cm} m{3cm}}
\toprule
\textbf{Step} & \textbf{Content} & \textbf{Evol Type} \\
\midrule
Seed Sample & Continuate a scrivere il seguente testo: Oliver ha colpito Bishop con tanta forza che il vecchio si e'stravolto all'indietro e e'caduto sul lato del trono; poi Oliver si e'spostato sul lato di Amelie. ``no!'' Oliver ha fatto un salto mentre Sam si aggrappava del palo per tirarlo fuori. ``lei e'vecchia.'' & \\
\midrule
Evolved (Turn 1) Instr. & C'era una volta un villaggio chiamato Lanciano, famoso per le sue feste. Un giorno, il più coraggioso tra gli abitanti, un uomo chiamato Oliver, decise di sfidare il vecchio e autoritario leader, Bishop. Così, durante la festa più importante dell'anno, Oliver \_\_\_\_ con tanta forza che il vecchio si e'stravolto all'indietro e e'caduto sul lato del palo delle bandiere. La folla si fermò, in attesa di vedere cosa sarebbe successo dopo. Poi Oliver si e'spostato sul lato di Amelie, la figlia di Bishop, e ha esclamato, ``Non è giusto! Lei è \_\_\_\_.'' I villaggi erano in shock. Sam, il migliore amico di Oliver, ha fatto un salto mentre si aggrappava al palo per \_\_\_\_. La tensione era palpabile. & Contextual Blanks \\
\midrule
GPT-4 Response & colpì Bishop\newline bellissima e merita di essere trattata con rispetto\newline stabilizzarsi & \\
\midrule
Turn 2 & Potresti riassumere la storia in punti elenco? & Format Rewriting Follow Up \\
\midrule
GPT-4 Response & Certamente! Ecco la storia riassunta in punti elenco:\newline\newline - Il villaggio di Lanciano è famoso per le sue feste.\newline - Oliver, il più coraggioso abitante del villaggio, decide di sfidare il vecchio e autoritario leader Bishop.\newline - Durante la festa più importante dell'anno, Oliver colpisce Bishop con tanta forza che quest'ultimo cade all'indietro, finendo vicino al palo delle bandiere.\newline - La folla si ferma, aspettando di vedere cosa accadrà dopo.\newline - Oliver si sposta accanto ad Amelie, la figlia di Bishop, e dichiara che non è giusto che lei venga trattata male perché è bellissima e merita rispetto.\newline - I villaggi sono in shock per la dichiarazione di Oliver.\newline - Sam, il migliore amico di Oliver, fa un salto e si aggrappa al palo per stabilizzarsi.\newline - La tensione nell'aria è palpabile. & \\
\midrule
Turn 3 & Per favore, riscrivi la storia in forma di elenco puntato. & Format Rewriting Follow Up \\
\midrule
GPT-4 Response & Certamente! Ecco la storia riscritta in forma di elenco puntato:\newline\newline - C'era una volta un villaggio chiamato Lanciano, famoso per le sue feste.\newline - Un giorno, Oliver, il più coraggioso tra gli abitanti, decise di sfidare il vecchio e autoritario leader, Bishop.\newline - Durante la festa più importante dell'anno, Oliver colpì Bishop con tanta forza che il vecchio si estraniò all'indietro e cadde vicino al palo delle bandiere.\newline - La folla si fermò, in attesa di vedere cosa sarebbe successo dopo.\newline - Oliver si spostò accanto ad Amelie, la figlia di Bishop, e esclamò: "Non è giusto! Lei è bellissima e merita di essere trattata con rispetto."\newline - I villaggi rimasero in shock.\newline - Sam, il migliore amico di Oliver, fece un salto e si aggrappò al palo per stabilizzarsi.\newline - La tensione era palpabile. & \\
\bottomrule
\end{tabular}
}
\caption{Conversation from \datasetname~in Italian.}
\label{tab:lanciano_story}
\end{table*}

\begin{table*}[h!]
\centering
\small
\resizebox{\textwidth}{!}{
\begin{tabular}{c m{10cm} m{3cm}}
\toprule
\textbf{Step} & \textbf{Content} & \textbf{Evol Type} \\
\midrule
Seed Sample & Este \'album, el segundo de un d\'uo notable en el \'ambito del hip hop, incluye colaboraciones de varios integrantes de un colectivo conocido por sus iniciales y su estilizaci\'on particular. Este grupo, que surgi\'o en la mitad de la primera d\'ecada del siglo XXI, est\'a asociado a una ciudad estadounidense. \textquestiondown Podr\'ias identificar dicha ciudad? & \\
\midrule
Evolved (Turn 1) Instr. & Este \'album, el segundo de un d\'uo notable en el \'ambito del hip hop, incluye colaboraciones de varios integrantes de un colectivo conocido por sus iniciales y su estilizaci\'on particular. Este grupo, que surgi\'o en la mitad de la primera d\'ecada del siglo XXI, est\'a asociado a una ciudad estadounidense. \textquestiondown Podr\'ias identificar dicha ciudad? & Ambiguity \\
\midrule
GPT-4 Response & Claro, parece que est\'as hablando del \'album ``Run the Jewels 2'' del d\'uo de hip hop Run the Jewels, compuesto por El-P y Killer Mike. Este \'album incluye colaboraciones de varios miembros del colectivo conocido como ``Odd Future'' (estilizado como ``OFWGKTA''). Odd Future es un colectivo de hip hop que surgi\'o a mediados de la d\'ecada de 2000 y est\'a asociado con la ciudad de Los \'Angeles, California. & \\
\midrule
Turn 2 & ``Ahora, por favor, identifica y enumera todas las canciones de ese \'album en particular, junto con sus respectivas duraciones y los artistas que colaboraron en cada una de ellas.'' & Challenging Follow-Up \\
\midrule
GPT-4 Response & El \'album ``Run the Jewels 2'' de Run the Jewels incluye las siguientes canciones:\newline\newline 1. ``Jeopardy'' - 3:21\newline 2. ``Oh My Darling Don\'t Cry'' - 3:24\newline 3. ``Blockbuster Night, Part 1'' - 2:32\newline 4. ``Close Your Eyes (And Count to Fuck)'' con Zack de la Rocha - 3:54\newline 5. ``All My Life'' - 3:08\newline 6. ``Lie, Cheat, Steal'' - 3:29\newline 7. ``Early'' con BOOTS - 3:44\newline 8. ``All Due Respect'' con Travis Barker - 3:40\newline 9. ``Love Again (Akinyele Back)'' con Gangsta Boo - 3:02\newline 10. ``Crown'' con Diane Coffee - 3:48\newline 11. ``Angel Duster'' - 5:12 & \\
\bottomrule
\end{tabular}
}
\caption{Conversation from \datasetname~in Spanish.}
\label{tab:spanish_text_example}
\end{table*}

\begin{table*}[h!]
\centering
\scriptsize
\resizebox{\textwidth}{!}{
\begin{tabular}{c m{10cm} m{3cm}}
\toprule
\textbf{Step} & \textbf{Content} & \textbf{Evol Type} \\
\midrule
Seed Sample & Bei Menschen kann ein Merkmal durch ein Paar oder viele Paare von was bestimmt werden? Wählen Sie aus den folgenden Optionen und begründen Sie Ihre Antwort detailliert:\newline 1. Gene\newline 2. Mikroben\newline 3. Zellen\newline 4. Organe & \\
\midrule
Evolved (Turn 1) Instr. & Bei Menschen kann ein Merkmal durch ein Paar oder viele Paare von was bestimmt werden? Wählen Sie aus den folgenden Optionen und begründen Sie Ihre Antwort detailliert:\newline 1. Gene\newline 2. Mikroben\newline 3. Zellen\newline 4. Organe & Justification \\
\midrule
GPT-4 Response & Das Merkmal bei Menschen kann durch ein Paar oder viele Paare von **Genen** bestimmt werden. Hier ist die detaillierte Begründung:\newline\newline **Gene**: Gene sind die grundlegenden Einheiten der Vererbung. Sie bestehen aus DNA und enthalten die Informationen, die für die Entwicklung, Funktion und Reproduktion von Organismen notwendig sind. Jedes Merkmal eines Menschen, wie Augenfarbe, Haarfarbe oder Blutgruppe, wird durch Gene bestimmt. Diese Gene befinden sich auf Chromosomen, die in Paaren vorkommen, wobei ein Chromosom von der Mutter und eines vom Vater vererbt wird. Viele Merkmale werden von mehreren Genen (polygene Vererbung) beeinflusst, die in komplexer Weise miteinander interagieren.\newline\newline **Mikroben**: Mikroben sind Mikroorganismen, wie Bakterien, Viren und Pilze. Obwohl Mikroben eine Rolle bei der Gesundheit und Krankheit von Menschen spielen können, bestimmen sie keine genetischen Merkmale eines Menschen. Sie können jedoch das Immunsystem beeinflussen und zur Entwicklung bestimmter Erkrankungen beitragen.\newline\newline **Zellen**: Zellen sind die grundlegenden Bausteine des Lebens. Alle Lebewesen bestehen aus Zellen, aber Zellen selbst bestimmen nicht die Merkmale eines Menschen. Die Funktionen und Eigenschaften der Zellen werden durch die in ihnen enthaltenen Gene reguliert.\newline\newline **Organe**: Organe sind Strukturen im Körper, die aus verschiedenen Geweben bestehen und spezifische Funktionen erfüllen, wie das Herz, die Leber oder die Nieren. Organe bestimmen keine genetischen Merkmale, sondern sind vielmehr das Ergebnis der Expression und Interaktion von Genen.\newline\newline Zusammengefasst, sind es die Gene, die die Merkmale eines Menschen durch ein oder mehrere Genpaare bestimmen. Diese Gene kodieren für Proteine und andere Moleküle, die die physischen und funktionalen Eigenschaften des Organismus beeinflussen. & \\
\midrule
Turn 2 & Könntest du bitte die Antwort kürzer und prägnanter formulieren? & Generic Rewriting Follow Up \\
\midrule
GPT-4 Response & Natürlich! Das Merkmal bei Menschen wird durch ein Paar oder viele Paare von **Genen** bestimmt. Gene bestehen aus DNA und enthalten die Informationen für die Entwicklung und Funktion von Organismen. Sie kommen in Paaren auf Chromosomen vor, die von beiden Elternteilen vererbt werden. Andere Optionen wie Mikroben, Zellen und Organe beeinflussen nicht direkt die genetischen Merkmale eines Menschen. & \\
\midrule
Turn 3 & Könnten Sie bitte näher erläutern, wie Gene das Merkmal eines Menschen bestimmen und welche Rolle genetische Variationen dabei spielen? & Expanding Queries \\
\midrule
GPT-4 Response & Gerne! Gene bestimmen die Merkmale eines Menschen durch die Kodierung von Proteinen und anderen Molekülen, die die physischen und funktionalen Eigenschaften des Organismus beeinflussen. Hier sind die wichtigsten Punkte dazu:\newline\newline \textbf{Funktionsweise der Gene}\newline 1. **DNA und Gene**: Gene bestehen aus DNA (Desoxyribonukleinsäure), die die Bauanleitung für Proteine enthält. Diese Proteine steuern viele Prozesse im Körper, wie Zellwachstum, Stoffwechsel und die Reaktion auf Umweltreize.\newline\newline 2. **Chromosomen**: Gene sind auf Chromosomen angeordnet. Menschen haben 23 Chromosomenpaare, wobei eines von jedem Elternteil kommt. Jedes Chromosom enthält viele Gene.\newline\newline 3. **Genexpression**: Gene werden durch Transkriptions- und Translationsprozesse in Proteine umgewandelt. Dieser Prozess wird als Genexpression bezeichnet und bestimmt, welche Proteine in welcher Menge produziert werden.\newline\newline \textbf{Genetische Variationen}\newline 1. **Allele**: Gene können in verschiedenen Varianten vorkommen, die als Allele bezeichnet werden. Ein Mensch hat für jedes Gen zwei Allele (eines von jedem Elternteil).\newline\newline 2. **Dominant und Rezessiv**: Manche Allele sind dominant und setzen sich gegenüber rezessiven Allelen durch. Ein dominantes Allel reicht aus, um ein bestimmtes Merkmal zu zeigen, während ein rezessives Allel nur dann zur Ausprägung kommt, wenn beide Allele rezessiv sind.\newline\newline 3. **Polygene Vererbung**: Viele Merkmale werden von mehreren Genen beeinflusst (polygene Vererbung). Zum Beispiel wird die Hautfarbe durch mehrere Gene bestimmt, die zusammenwirken.\newline\newline 4. **Mutation**: Mutationen sind Veränderungen in der DNA-Sequenz eines Gens. Sie können neue Allele erzeugen und zu genetischer Variation beitragen.\newline\newline \textbf{Rolle der genetischen Variation}\newline 1. **Einzigartigkeit**: Genetische Variationen führen dazu, dass jeder Mensch ein einzigartiges genetisches Profil hat. Diese Variationen beeinflussen Merkmale wie Aussehen, Anfälligkeit für Krankheiten und viele andere Eigenschaften.\newline\newline 2. **Anpassungsfähigkeit**: Genetische Diversität ist wichtig für die Anpassungsfähigkeit einer Population an sich ändernde Umweltbedingungen. Sie ermöglicht die natürliche Selektion und Evolution.\newline\newline Zusammengefasst, bestimmen Gene die Merkmale eines Menschen durch die Kodierung von Proteinen, die die biologischen Funktionen steuern. Genetische Variationen, wie unterschiedliche Allele und Mutationen, tragen zur Vielfalt der menschlichen Merkmale bei und beeinflussen, wie diese Merkmale ausgeprägt werden. & \\
\bottomrule
\end{tabular}
}
\caption{Conversation from \datasetname~in German.}
\label{tab:german_text_example}
\end{table*}

\begin{table*}[h!]
\centering
\small
\resizebox{\textwidth}{!}{
\begin{tabular}{c m{10cm} m{3cm}}
\toprule
\textbf{Step} & \textbf{Content} & \textbf{Evol Type} \\
\midrule
Seed Sample & Répondez à cette question: On mesure un tapis neuf pour une pièce de 5,5 m sur 4,6 m. Sur un ruban à mesurer, quelle est la plus petite unité nécessaire pour mesurer la pièce avec précision pour le tapis? Sélectionnez parmi: (1) en millièmes. (2) centimètres. (3) à des mètres. (4) kilomètres. (5) millimètres. (6) décamètres. Réponse: & \\
\midrule
Evolved (Turn 1) Instr. & Répondez à cette question: On mesure un tapis neuf pour une pièce de 5,5 m sur 4,6 m. Sur un ruban à mesurer, quelle est la plus petite unité nécessaire pour mesurer la pièce avec précision pour le tapis? Sélectionnez parmi: (1) en millièmes. (2) centimètres. (3) à des mètres. (4) kilomètres. (5) millimètres. (6) décamètres. Réponse: & More Choices \\
\midrule
GPT-4 Response & Pour mesurer une pièce de 5,5 m sur 4,6 m avec précision pour un tapis, la plus petite unité nécessaire serait le **millimètre**. Les millimètres permettent une mesure très précise qui est souvent nécessaire pour des ajustements fins dans des projets comme l'installation de tapis.\newline\newline Ainsi, la réponse est : **(5) millimètres**. & \\
\midrule
Turn 2 & Maintenant, reformulez votre réponse sous forme de liste à puces. & Format Rewriting Follow-Up \\
\midrule
GPT-4 Response & - Pour mesurer une pièce de 5,5 m sur 4,6 m avec précision pour un tapis, la plus petite unité nécessaire serait le **millimètre**.\newline - Les millimètres permettent une mesure très précise qui est souvent nécessaire pour des ajustements fins dans des projets comme l'installation de tapis.\newline\newline Ainsi, la réponse est :\newline - **(5) millimètres**. & \\
\midrule
Turn 3 & Pouvez-vous expliquer pourquoi vous avez choisi cette unité de mesure en particulier pour mesurer la pièce avec précision pour le tapis ? & Clarification \\
\midrule
GPT-4 Response & Bien sûr, voici une explication détaillée :\newline\newline - **Précision nécessaire pour le tapis** : Lors de l'installation d'un tapis, il est crucial de mesurer la pièce avec une grande précision pour s'assurer que le tapis s'adapte parfaitement sans laisser de vides ni nécessiter de grandes découpes supplémentaires.\newline\newline - **Millimètres versus autres unités** :\newline - **Millimètres** : Ils offrent la plus grande précision parmi les options données. Chaque millimètre étant un millième de mètre, il permet de capturer les dimensions exactes nécessaires pour un ajustement parfait.\newline - **Centimètres** : Bien que précis, les centimètres (1 cm = 10 mm) ne sont pas aussi précis que les millimètres et pourraient omettre des détails fins nécessaires à une coupe parfaite.\newline - **Mètres** : Mesurer en mètres (1 m = 1000 mm) serait trop grossier pour des dimensions aussi précises et fines.\newline - **Décamètres** et **kilomètres** : Ces unités sont beaucoup trop grandes pour une tâche de cette envergure et ne sont pas pratiques pour des mesures de précision dans un contexte domestique.\newline - **Millièmes** : S'il s'agit de millièmes de mètre, cela serait encore plus précis que nécessaire, et pour des mesures domestiques, les millimètres suffisent largement.\newline\newline En conclusion, les **millimètres** sont l'unité de mesure la plus appropriée pour garantir que le tapis est coupé et installé avec une précision maximale, assurant ainsi un ajustement parfait. & \\
\bottomrule
\end{tabular}
}
\caption{Conversation from \datasetname~in French.}
\label{tab:french_text_example}
\end{table*}

\subsection{Prompt Taxonomy for Evol-instruct}
\label{sec:evol_prompts}
For Dolly, HotpotQA and MLQA we use evols from generic, OpenQA and Mintaka respectively.
\begin{table}[htbp!]
    \centering
    \scriptsize
    \begin{tabular}{>{\raggedright}p{0.2\textwidth} >{\raggedright\arraybackslash}p{0.75\textwidth}}
    \toprule
    \textbf{Evol Type} & \textbf{GPT-4 Prompt} \\
    \midrule
    Transliteration & Rewrite the \#given\_prompt\# as an <translit\_language> transliteration, and create \#new\_prompt\#. Additionally, conclude with a request to respond in <translit\_language> transliteration.
\#given\_prompt\#:
<prompt> \\
    \midrule
    Dialect & You are a brilliant <prompt\_language> native speaker. Rewrite \#given\_prompt\# by changing the dialect to <prompt\_language> and create \#new\_prompt\#. Finally, ask to respond in the same <prompt\_language> dialect. Write the \#new\_prompt\# prompt in <prompt\_language>.
\#given\_prompt\#:
<prompt> \\
    \midrule
    Concise & Re-write the \#given\_prompt\# concisely, and create \#new\_prompt\#. Additionally, conclude with a request to respond concisely.  Write the \#new\_prompt\# in <prompt\_language>.
\#given\_prompt\#:
<prompt> \\
    \midrule
    Deepen & Slightly increase the depth and breadth of \#given\_prompt\#, and create \#new\_prompt\#. Write the \#new\_prompt\# in <prompt\_language>.
\#given\_prompt\#:
<prompt> \\
    \midrule
    Concretize & Make \#given\_prompt\# slightly more concrete, and create \#new\_prompt\#. Additionally, conclude with a request for an AI assistant to respond with the a detailed and concrete response. Write the \#new\_prompt\# in <prompt\_language>.
\#given\_prompt\#:
<prompt> \\
    \midrule
    Increase Reasoning & If \#given\_prompt\# can be solved with just a few simple thinking processes, rewrite it to explicitly request multi-step reasoning, and create \#new\_prompt\#. Write the \#new\_prompt\# in <prompt\_language>.
\#given\_prompt\#:
<prompt> \\
        \bottomrule
    \end{tabular}
    \caption{Generic}
    \label{tab:generic}
\end{table}

\begin{table}[htbp!]
    \centering
    \scriptsize
    \begin{tabular}{>{\raggedright}p{0.2\textwidth} >{\raggedright\arraybackslash}p{0.75\textwidth}}
    \toprule
    \textbf{Evol Type} & \textbf{GPT-4 Prompt} \\
    \midrule
    Adding Distractors & Given a prompt \#given\_prompt\# that represents an article to be summarized, based upon the \#given\_prompt\# create a \#new\_prompt\# in the same language that adds unrelated or distracting information in the article which is not relevant to the main topic.
\#given\_prompt\#:
<prompt> \\
    \midrule
    Technical Jargon & Given a prompt \#given\_prompt\# that represents an article to be summarized, based upon the \#given\_prompt\# create a \#new\_prompt\# in the same language by adding technical jargon or industry specifc terms that makes it difficult to summarize.
\#given\_prompt\#:
<prompt> \\
    \midrule
    Inconsistencies In Information & Given a prompt \#given\_prompt\# that represents an article to be summarized, based upon the \#given\_prompt\# create a \#new\_prompt\# in the same language to include contradictions or inconsistencies within the article thus forcing the summarizer to discern which piece of information is accurate and relevant.
\#given\_prompt\#:
<prompt> \\
    \midrule
    Multiple Topics & Given a prompt \#given\_prompt\# that represents an article to be summarized, based upon the \#given\_prompt\# create a \#new\_prompt\# in the same language that covers multiple topics or subtopics thus make summarization more complicated.
\#given\_prompt\#:
<prompt> \\
    \midrule
    Metaphors Idiom & Given a prompt \#given\_prompt\# that represents an article to be summarized, based upon the \#given\_prompt\# create a \#new\_prompt\# using complex metaphors, idioms and cultural references thus making summarization more challenging.
\#given\_prompt\#:
<prompt> \\
    \midrule
    Long Distance & Given a prompt \#given\_prompt\# that represents an article to be summarized, based upon the \#given\_prompt\# create a \#new\_prompt\# by increasing the distance between related pieces of information in the text as this requires understanding the deeper structure of the text.
\#given\_prompt\#:
<prompt> \\
    \midrule
    Multiple Languages & Given a prompt \#given\_prompt\# that represents an article to be summarized, based upon the \#given\_prompt\# create a \#new\_prompt\# in <language\_2> and asking to respond in <language\_1> thus making the task challenging as it requires understanding and proficiency in more than one language.
\#given\_prompt\#:
<prompt> \\
    \midrule
    Unstructured Data & Given a prompt \#given\_prompt\# that represents an article to be summarized, based upon the \#given\_prompt\# create a \#new\_prompt\# by presenting the article in a non-linear or non-chronological format thus increasing the complexity as it becomes challenging to pick out the main points and summarize them accurately.
\#given\_prompt\#:
<prompt> \\
    \midrule
    Personal Opinion & Given a prompt \#given\_prompt\# that represents an article to be summarized, based upon the \#given\_prompt\# create a \#new\_prompt\# by incorporating bias or personal opinion as it greatly complicates the summarization process as the summarizer needs to remain neutral and objective.
\#given\_prompt\#:
<prompt> \\
        \bottomrule
    \end{tabular}
    \caption{Abstract Summarization}
    \label{tab:abstract_summarization}
\end{table}

\begin{table}[htbp!]
    \centering
    \scriptsize
    \begin{tabular}{>{\raggedright}p{0.2\textwidth} >{\raggedright\arraybackslash}p{0.75\textwidth}}
    \toprule
    \textbf{Evol Type} & \textbf{GPT-4 Prompt} \\
    \midrule
    Jargon & Given a prompt \#given\_prompt\# that asks an explanation of a joke, based upon the \#given\_prompt\# create a \#new\_prompt\# in the same language by incorporating jargon to the jokes that are specific to a certain profession, field, or hobby thus requiring deeper knowledge of the field in order to explain the joke properly.
\#given\_prompt\#:
<prompt> \\
    \midrule
    Slang & Given a prompt \#given\_prompt\# that asks an explanation of a joke, based upon the \#given\_prompt\# create a \#new\_prompt\# in the same language that use slang or colloquial language, thus making it harder to understand and explain the punchline.
\#given\_prompt\#:
<prompt> \\
    \midrule
    Cultural References & Given a prompt \#given\_prompt\# that asks an explanation of a joke, based upon the \#given\_prompt\# create a \#new\_prompt\# in the same language that increasingly use jokes which are culture-specific as will require cultural understanding tp provide explanations.
\#given\_prompt\#:
<prompt> \\
    \midrule
    Non Explicit Punchline & Given a prompt \#given\_prompt\# that asks an explanation of a joke, based upon the \#given\_prompt\# create a \#new\_prompt\# in the same language such that jokes have a punchline isn't explicitly stated, but rather implied.
\#given\_prompt\#:
<prompt> \\
    \midrule
    Time Bound & Given a prompt \#given\_prompt\# that represents an article and a question related to that article, based upon the \#given\_prompt\# create a \#new\_prompt\# by making jokes that were relative to a certain time period or current event, thus making it harder to grasp.
\#given\_prompt\#:
<prompt> \\
    \midrule
    Compound Jokes & Given a prompt \#given\_prompt\# that asks an explanation of a joke, based upon the \#given\_prompt\# create a \#new\_prompt\# in the same language such that it uses compound jokes which contain multiple punchlines within the same joke. This would make the explanation task difficult as one would need to explain multiple punchlines coherently.
\#given\_prompt\#:
<prompt> \\
    \midrule
    Language & Given a prompt \#given\_prompt\# that asks an explanation of a joke, based upon the \#given\_prompt\# create a \#new\_prompt\# in <language\_2> such that it challenge the language skills. This would make the explanation task difficult as one would need to explain multiple punchlines coherently. Finally ask to respond with explanation in <language\_1>
\#given\_prompt\#:
<prompt> \\
    \midrule
    Long & Given a prompt \#given\_prompt\# that asks an explanation of a joke, based upon the \#given\_prompt\# create a \#new\_prompt\# in the same language but make the jokes exceedingly long where the punchline isn't delivered immediately and requires you to remember or understand preceding parts of the joke.
\#given\_prompt\#:
<prompt> \\
    \midrule
    Double Entendre & Given a prompt \#given\_prompt\# that asks an explanation of a joke, based upon the \#given\_prompt\# create a \#new\_prompt\# in the same language but utilize jokes with double entendre, where there are two possible interpretations.
\#given\_prompt\#:
<prompt> \\
        \bottomrule
    \end{tabular}
    \caption{Joke Explain}
    \label{tab:joke_explain}
\end{table}

\begin{table}[htbp!]
    \centering
    \scriptsize
    \begin{tabular}{>{\raggedright}p{0.2\textwidth} >{\raggedright\arraybackslash}p{0.75\textwidth}}
    \toprule
    \textbf{Evol Type} & \textbf{GPT-4 Prompt} \\
    \midrule
    More Choices & Given a prompt \#given\_prompt\# that represents a topic and multiple choice question related to that topic, based upon the \#given\_prompt\# create a \#new\_prompt\# by changing adding more choices to the question that are relevant to the topic but not correct. This will make it challenging to answer.
\#given\_prompt\#:
<prompt> \\
    \midrule
    More Complex & Given a prompt \#given\_prompt\# that represents a topic and multiple choice question related to that topic, based upon the \#given\_prompt\# create a \#new\_prompt\# by making it more complex using technical or domain specific jargon This will make it challenging to understand the quesrion thus making it difficult to answer.
\#given\_prompt\#:
<prompt> \\
    \midrule
    Negation & Given a prompt \#given\_prompt\# that represents a topic and multiple choice question related to that topic, based upon the \#given\_prompt\# create a \#new\_prompt\# by by asking negative questions that require the recognition of the negation included in the sentences.
\#given\_prompt\#:
<prompt> \\
    \midrule
    Multiple Corrct Options & Given a prompt \#given\_prompt\# that represents a topic and multiple choice question related to that topic, based upon the \#given\_prompt\# create a \#new\_prompt\# by changing adding more choices to the question that are correct and relevant to the topic. This will make it challenging as one will need to choose all correct options.
\#given\_prompt\#:
<prompt> \\
    \midrule
    Need For Context & Given a prompt \#given\_prompt\# that represents a topic and multiple choice question related to that topic, based upon the \#given\_prompt\# create a \#new\_prompt\# by asking questions that require additional context other than the one provided in the topic.This will make it challenging as it will evaluate the knowledge someone has on the topic.
\#given\_prompt\#:
<prompt> \\
    \midrule
    Justification & Given a prompt \#given\_prompt\# that represents a topic and multiple choice question related to that topic, based upon the \#given\_prompt\# create a \#new\_prompt\# by asking to respond with the correct answer and provide a detailed justification for the answer.
\#given\_prompt\#:
<prompt> \\
    \midrule
    Distracting Long & Given a prompt \#given\_prompt\# that represents a topic and multiple choice question related to that topic, based upon the \#given\_prompt\# create a \#new\_prompt\# by asking questions that require additional context other than the one provided in the topic. This will make it challenging as it will evaluate the knowledge someone has on the topic.
\#given\_prompt\#:
<prompt> \\
    \midrule
    Close Choices & Given a prompt \#given\_prompt\# that represents a topic and multiple choice question related to that topic, based upon the \#given\_prompt\# create a \#new\_prompt\# by addding more choices to the question that are closely related to each other. This will make the task more challenging.
\#given\_prompt\#:
<prompt> \\
    \midrule
    No Correct Option & Given a prompt \#given\_prompt\# that represents a topic and multiple choice question related to that topic, based upon the \#given\_prompt\# create a \#new\_prompt\# by changing the choices such that no choice is the correct answer.
\#given\_prompt\#:
<prompt> \\
        \bottomrule
    \end{tabular}
    \caption{Flan Qa}
    \label{tab:flan_qa}
\end{table}

\begin{table}[htbp!]
    \centering
    \scriptsize
    \begin{tabular}{>{\raggedright}p{0.2\textwidth} >{\raggedright\arraybackslash}p{0.75\textwidth}}
    \toprule
    \textbf{Evol Type} & \textbf{GPT-4 Prompt} \\
    \midrule
    Complex Jargon & Given a prompt \#given\_prompt\# that requires answering a question about a text snippet, based upon the \#given\_prompt\# create a \#new\_prompt\# in the same language by either using technical jargons, domain-specific language, technical or scientific complexities thus making the task more challenging as it requires deep understanding and specialized knowledge to answer.
\#given\_prompt\#:
<prompt> \\
    \midrule
    Multiple Languages & Given a prompt \#given\_prompt\# that requires answering a question about a text snippet, based upon the \#given\_prompt\# create a \#new\_prompt\# in <language\_2> including code-switching i.e. switching between languages within a single conversation or sentence. Finally ask to respond in <LANGUAG\_1>
\#given\_prompt\#:
<prompt> \\
    \midrule
    Long Complex Queries & Given a prompt \#given\_prompt\# that requires answering a question about a text snippet, based upon the \#given\_prompt\# create a \#new\_prompt\# in the same language by providing vital pieces of information in the text snippet in a non-linear, disconnected manner thus requiring piecing them together accurately to form an explanation..
\#given\_prompt\#:
<prompt> \\
    \midrule
    Disconnected Clues & Given a prompt \#given\_prompt\# that requires answering a question about a text snippet, based upon the \#given\_prompt\# create a \#new\_prompt\# in the same language by translating either the question or the text snippet (but not both) in any language thus making the task more challenging as it requires understanding or different languages.
\#given\_prompt\#:
<prompt> \\
    \midrule
    Emotion Sarcasam & Given a prompt \#given\_prompt\# that requires answering a question about a text snippet, based upon the \#given\_prompt\# create a \#new\_prompt\# in the same language by adding emotion or sarcasm in the text snippet as recognizing and responding can be a huge challenge.
\#given\_prompt\#:
<prompt> \\
    \midrule
    Advanced Reasoning & Given a prompt \#given\_prompt\# that represents two sentences and asks whether the two are paraphrases or not, based upon the \#given\_prompt\# create a \#new\_prompt\# in the same language by writing text snippet or question that require advanced logic or reasoning, such as those found in certain categories of IQ test, thus make it more difficult to answer.
\#given\_prompt\#:
<prompt> \\
    \midrule
    Ambiguity & Given a prompt \#given\_prompt\# that represents two sentences and asks whether the two are paraphrases or not, based upon the \#given\_prompt\# create a \#new\_prompt\# in the same language by including deceptive or ambiguous phrases that might lead to misinterpretation can complicate the task.
\#given\_prompt\#:
<prompt> \\
    \midrule
    Unknown Context & Given a prompt \#given\_prompt\# that represents two sentences and asks whether the two are paraphrases or not, based upon the \#given\_prompt\# create a \#new\_prompt\# in the same language by adding larger context not included in the text, as it would require to infer missing details, which adds complexity to the task.
\#given\_prompt\#:
<prompt> \\
    \midrule
    Implicit Bias & Given a prompt \#given\_prompt\# that represents two sentences and asks whether the two are paraphrases or not, based upon the \#given\_prompt\# create a \#new\_prompt\# in the same language by including subtle biases or nuances, recognizing and appropriately responding to these can be challenging.
\#given\_prompt\#:
<prompt> \\
        \bottomrule
    \end{tabular}
    \caption{Flan Cot}
    \label{tab:flan_cot}
\end{table}

\begin{table}[htbp!]
    \centering
    \scriptsize
    \begin{tabular}{>{\raggedright}p{0.2\textwidth} >{\raggedright\arraybackslash}p{0.75\textwidth}}
    \toprule
    \textbf{Evol Type} & \textbf{GPT-4 Prompt} \\
    \midrule
    Ambiguity & Given a prompt \#given\_prompt\# that represents an article and multiple questions related to that article, based upon the \#given\_prompt\# create a \#new\_prompt\# by changing the questions only in the same language but making it much more vague and ambiguous thus making it not so straightforward to answer.
\#given\_prompt\#:
<prompt> \\
    \midrule
    Long Form Question & Given a prompt \#given\_prompt\# that represents an article and multiple questions related to that article, based upon the \#given\_prompt\# create a \#new\_prompt\# by changing the questions only in the same language by making it longer i.e. formulating the questions in long and complex sentences thus requiring the system to decipher the main question.
\#given\_prompt\#:
<prompt> \\
    \midrule
    Multilingual & Given a prompt \#given\_prompt\# that represents an article and multiple questions related to that article, based upon the \#given\_prompt\# create a \#new\_prompt\# by changing the questions in <language\_2> and article in <language\_1>, having different linguistic structure. Finally, ask to answer the question in the <language\_2>.
\#given\_prompt\#:
<prompt> \\
    \midrule
    Combine Facts & Given a prompt \#given\_prompt\# that represents an article and multiple questions related to that article, based upon the \#given\_prompt\# create a \#new\_prompt\# by changing the questions only in the same language by combining multiple facts thus making the questions more complex and requiring combining multiple facts to answer correctly.
\#given\_prompt\#:
<prompt> \\
    \midrule
    Implicit Question & Given a prompt \#given\_prompt\# that represents an article and multiple questions related to that article, based upon the \#given\_prompt\# create a \#new\_prompt\# by changing the questions only in the same language by asking implicit questions where the answer isn't explicit and requires understanding of the underlying implication.
\#given\_prompt\#:
<prompt> \\
    \midrule
    Negative Questions & Given a prompt \#given\_prompt\# that represents an article and multiple questions related to that article, based upon the \#given\_prompt\# create a \#new\_prompt\# by asking negative questions in the same language that require the recognition of the negation included in the sentences.
\#given\_prompt\#:
<prompt> \\
    \midrule
    Inference & Given a prompt \#given\_prompt\# that represents an article and multiple questions related to that article, based upon the \#given\_prompt\# create a \#new\_prompt\# by changing the question only in the same language by asking questions that require a degree of inference or deduction not directly provided.
\#given\_prompt\#:
<prompt> \\
    \midrule
    Multichoice Questions & Given a prompt \#given\_prompt\# that represents an article and multiple questions related to that article, based upon the \#given\_prompt\# create a \#new\_prompt\# by asking multiple choice questions in the same language based upon the given article, where the choices are from the article itself. Finally ask the model to respond with the correct choice and explain the decision.
\#given\_prompt\#:
<prompt> \\
    \midrule
    More Reasoning & Given a prompt \#given\_prompt\# that represents an article and multiple questions related to that article, based upon the \#given\_prompt\# create a \#new\_prompt\# by asking questions in the same language that require multistep reasoning processes, where participants need to follow a sequence of logical steps to arrive at the correct answe.
\#given\_prompt\#:
<prompt> \\
        \bottomrule
    \end{tabular}
    \caption{Flan Coqa}
    \label{tab:flan_coqa}
\end{table}

\begin{table}[htbp!]
    \centering
    \scriptsize
    \begin{tabular}{>{\raggedright}p{0.2\textwidth} >{\raggedright\arraybackslash}p{0.75\textwidth}}
    \toprule
    \textbf{Evol Type} & \textbf{GPT-4 Prompt} \\
    \midrule
    Multiple Blanks & Given a prompt \#given\_prompt\# that asks to either complete the sentence with a phrase or fill in the blank in between the text, based upon the \#given\_prompt\# create a \#new\_prompt\# in the same language but, that requires multiple blanks to be filled in.
\#given\_prompt\#:
<prompt> \\
    \midrule
    Contextual Blanks & Given a prompt \#given\_prompt\# that asks to either complete the sentence with a phrase or fill in the blank in between the text, based upon the \#given\_prompt\# create a \#new\_prompt\# in the same language but, that includes contextual blanks within a paragraph or story.
\#given\_prompt\#:
<prompt> \\
    \midrule
    Word Bank & Given a prompt \#given\_prompt\# that asks to either complete the sentence with a phrase or fill in the blank in between the text, based upon the \#given\_prompt\# create a \#new\_prompt\# in the same language but, that provides a word bank with distractors for filling in the blank.
\#given\_prompt\#:
<prompt> \\
    \midrule
    Missing Letters & Given a prompt \#given\_prompt\# that asks to either complete the sentence with a phrase or fill in the blank in between the text, based upon the \#given\_prompt\# create a \#new\_prompt\# in the same language but, that includes sentences with missing letters to be filled in.
\#given\_prompt\#:
<prompt> \\
    \midrule
    Grammar Forms & Given a prompt \#given\_prompt\# that asks to either complete the sentence with a phrase or fill in the blank in between the text, based upon the \#given\_prompt\# create a \#new\_prompt\# in the same language but, that requires choosing the correct grammatical form of a word for the blank.
\#given\_prompt\#:
<prompt> \\
    \midrule
    Logical Inferences & Given a prompt \#given\_prompt\# that asks to either complete the sentence with a phrase or fill in the blank in between the text, based upon the \#given\_prompt\# create a \#new\_prompt\# in the same language but, that requires logical inference for filling in the blank.
\#given\_prompt\#:
<prompt> \\
    \midrule
    Word Classes & Given a prompt \#given\_prompt\# that asks to either complete the sentence with a phrase or fill in the blank in between the text, based upon the \#given\_prompt\# create a \#new\_prompt\# in the same language but, that specifies the type of word needed for the blank.
\#given\_prompt\#:
<prompt> \\
    \midrule
    Synonyms Antonyms & Given a prompt \#given\_prompt\# that asks to either complete the sentence with a phrase or fill in the blank in between the text, based upon the \#given\_prompt\# create a \#new\_prompt\# in the same language but, that requires choosing a synonym or antonym for the blank.
\#given\_prompt\#:
<prompt> \\
    \midrule
    Cultural Context & Given a prompt \#given\_prompt\# that asks to either complete the sentence with a phrase or fill in the blank in between the text, based upon the \#given\_prompt\# create a \#new\_prompt\# in the same language but, that includes cultural references or idiomatic expressions.
\#given\_prompt\#:
<prompt> \\
        \bottomrule
    \end{tabular}
    \caption{Flan Lambda}
    \label{tab:flan_lambda}
\end{table}

\begin{table}[htbp!]
    \centering
    \scriptsize
    \begin{tabular}{>{\raggedright}p{0.2\textwidth} >{\raggedright\arraybackslash}p{0.75\textwidth}}
    \toprule
    \textbf{Evol Type} & \textbf{GPT-4 Prompt} \\
    \midrule
    Cross Lingual & Given a prompt \#given\_prompt\# that asks to generate a question for a certain text snippet or a topic, based upon the \#given\_prompt\# create a \#new\_prompt\# in <language\_2> and also asking to respond in <language\_1>.
\#given\_prompt\#:
<prompt> \\
    \midrule
    Domain Language & Given a prompt \#given\_prompt\# that asks to generate a question for a certain text snippet or a topic, based upon the \#given\_prompt\# create a \#new\_prompt\# in the same language by introducing domain specific language and related to the specialized field or topic described in the artic;e.
\#given\_prompt\#:
<prompt> \\
    \midrule
    Emotional Subtext & Given a prompt \#given\_prompt\# that asks to generate a question for a certain text snippet or a topic, based upon the \#given\_prompt\# create a \#new\_prompt\# in the same language by including sarcasm, euphemism, or other nuanced forms of communication thus make it harder to determine the possible question for the topic.
\#given\_prompt\#:
<prompt> \\
    \midrule
    Need For Context & Given a prompt \#given\_prompt\# that asks to generate a question for a certain text snippet or a topic, based upon the \#given\_prompt\# create a \#new\_prompt\# in the same language that requires additional context or background to determine the relevant question for the text snippet.
\#given\_prompt\#:
<prompt> \\
    \midrule
    Ambiguity In Wording & Given a prompt \#given\_prompt\# that asks to generate a question for a certain text snippet or a topic, based upon the \#given\_prompt\# create a \#new\_prompt\# in the same language by adding ambiguity to the text snippet thus making it more challenging to come up with a relevant question.
\#given\_prompt\#:
<prompt> \\
    \midrule
    Long Text & Given a prompt \#given\_prompt\# that asks to generate a question for a certain text snippet or a topic, based upon the \#given\_prompt\# create a \#new\_prompt\# in the same language by detailing and make the text snippet much longer thus making it more challenging to come up with a relevant question.
\#given\_prompt\#:
<prompt> \\
    \midrule
    Idiom Slang & Given a prompt \#given\_prompt\# that asks to generate a question for a certain text snippet or a topic, based upon the \#given\_prompt\# create a \#new\_prompt\# by the use of idiomatic expressions or regional slang thus obscuring the meaning of text snippet and making it more challenging to come up with a question.
\#given\_prompt\#:
<prompt> \\
    \midrule
    Adding Distractors & Given a prompt \#given\_prompt\# that represents an article to be summarized, based upon the \#given\_prompt\# create a \#new\_prompt\# in the same language that adds unrelated or distracting information in the article which is not relevant to the main topic.
\#given\_prompt\#:
<prompt> \\
    \midrule
    Multiple Questions & Given a prompt \#given\_prompt\# that represents an article to be summarized, based upon the \#given\_prompt\# create a \#new\_prompt\# in the same language by asking to come up with multiple questions regarding the text snippet.
\#given\_prompt\#:
<prompt> \\
        \bottomrule
    \end{tabular}
    \caption{Answer Ranking}
    \label{tab:answer_ranking}
\end{table}

\begin{table}[htbp!]
    \centering
    \scriptsize
    \begin{tabular}{>{\raggedright}p{0.2\textwidth} >{\raggedright\arraybackslash}p{0.75\textwidth}}
    \toprule
    \textbf{Evol Type} & \textbf{GPT-4 Prompt} \\
    \midrule
    Ambiguity & Given a prompt \#given\_prompt\# that represents an article and a question related to that article, based upon the \#given\_prompt\# create a \#new\_prompt\# by changing the question only in the same language but making it much more vague and ambiguous thus making it not so straightforward to answer.
\#given\_prompt\#:
<prompt> \\
    \midrule
    Long Form Question & Given a prompt \#given\_prompt\# that represents an article and a question related to that article, based upon the \#given\_prompt\# create a \#new\_prompt\# by changing the question only in the same language by making it longer i.e. formulating the questions in long and complex sentences thus requiring the system to decipher the main question.
\#given\_prompt\#:
<prompt> \\
    \midrule
    Multilingual & Given a prompt \#given\_prompt\# that represents an article and a question related to that article, based upon the \#given\_prompt\# create a \#new\_prompt\# by changing the question in <language\_2> and article in <language\_1>, having different linguistic structure. Finally, ask to answer the question in the <language\_2>.
\#given\_prompt\#:
<prompt> \\
    \midrule
    Combine Facts & Given a prompt \#given\_prompt\# that represents an article and a question related to that article, based upon the \#given\_prompt\# create a \#new\_prompt\# by changing the question only in the same language by combining multiple facts thus making the question more complex and requiring combining multiple facts to answer correctly.
\#given\_prompt\#:
<prompt> \\
    \midrule
    Implicit Question & Given a prompt \#given\_prompt\# that represents an article and a question related to that article, based upon the \#given\_prompt\# create a \#new\_prompt\# by changing the question only in the same language by asking implicit question where the answer isn't explicit and requires understanding of the underlying implication.
\#given\_prompt\#:
<prompt> \\
    \midrule
    Negative Questions & Given a prompt \#given\_prompt\# that represents an article and a question related to that article, based upon the \#given\_prompt\# create a \#new\_prompt\# by changing the question only in the same language by asking negative questions that require the recognition of the negation included in the sentences.
\#given\_prompt\#:
<prompt> \\
    \midrule
    Inference & Given a prompt \#given\_prompt\# that represents an article and a question related to that article, based upon the \#given\_prompt\# create a \#new\_prompt\# by changing the question only in the same language by asking questions that require a degree of inference or deduction not directly provided.
\#given\_prompt\#:
<prompt> \\
    \midrule
    Multichoice Questions & Given a prompt \#given\_prompt\# that represents an article and a question related to that article, based upon the \#given\_prompt\# create a \#new\_prompt\# by asking multiple choice question based upon the given article, where the choices are from the article itself. Finally ask the model to respond with the correct choice and explain the decision.
\#given\_prompt\#:
<prompt> \\
    \midrule
    More Reasoning & Given a prompt \#given\_prompt\# that represents an article and a question related to that article, based upon the \#given\_prompt\# create a \#new\_prompt\# by asking  that require multistep reasoning processes, where participants need to follow a sequence of logical steps to arrive at the correct answe.
\#given\_prompt\#:
<prompt> \\
        \bottomrule
    \end{tabular}
    \caption{Mintaka}
    \label{tab:mintaka}
\end{table}

\begin{table}[htbp!]
    \centering
    \scriptsize
    \begin{tabular}{>{\raggedright}p{0.2\textwidth} >{\raggedright\arraybackslash}p{0.75\textwidth}}
    \toprule
    \textbf{Evol Type} & \textbf{GPT-4 Prompt} \\
    \midrule
    Adding Distractors & Given a prompt \#given\_prompt\# that represents an article to be summarized, based upon the \#given\_prompt\# create a \#new\_prompt\# in the same language that adds unrelated or distracting information in the article which is not relevant to the main topic.
\#given\_prompt\#:
<prompt> \\
    \midrule
    Technical Jargon & Given a prompt \#given\_prompt\# that represents an article to be summarized, based upon the \#given\_prompt\# create a \#new\_prompt\# in the same language by adding technical jargon or industry specifc terms that makes it difficult to summarize.
\#given\_prompt\#:
<prompt> \\
    \midrule
    Inconsistencies In Information & Given a prompt \#given\_prompt\# that represents an article to be summarized, based upon the \#given\_prompt\# create a \#new\_prompt\# in the same language to include contradictions or inconsistencies within the article thus forcing the summarizer to discern which piece of information is accurate and relevant.
\#given\_prompt\#:
<prompt> \\
    \midrule
    Multiple Topics & Given a prompt \#given\_prompt\# that represents an article to be summarized, based upon the \#given\_prompt\# create a \#new\_prompt\# in the same language that covers multiple topics or subtopics thus make summarization more complicated.
\#given\_prompt\#:
<prompt> \\
    \midrule
    Metaphors Idiom & Given a prompt \#given\_prompt\# that represents an article to be summarized, based upon the \#given\_prompt\# create a \#new\_prompt\# using complex metaphors, idioms and cultural references thus making summarization more challenging.
\#given\_prompt\#:
<prompt> \\
    \midrule
    Long Distance & Given a prompt \#given\_prompt\# that represents an article to be summarized, based upon the \#given\_prompt\# create a \#new\_prompt\# by increasing the distance between related pieces of information in the text as this requires understanding the deeper structure of the text.
\#given\_prompt\#:
<prompt> \\
    \midrule
    Cross Lingual Summary & Given a prompt \#given\_prompt\# that represents some article about a topic, rewrite the article only and create a \#new\_prompt\# in any <language\_2>. Finally ask to provide a summary in the <language\_1>.
\#given\_prompt\#:
<prompt> \\
    \midrule
    Unstructured Data & Given a prompt \#given\_prompt\# that represents an article to be summarized, based upon the \#given\_prompt\# create a \#new\_prompt\# by presenting the article in a non-linear or non-chronological format thus increasing the complexity as it becomes challenging to pick out the main points and summarize them accurately.
\#given\_prompt\#:
<prompt> \\
    \midrule
    Personal Opinion & Given a prompt \#given\_prompt\# that represents an article to be summarized, based upon the \#given\_prompt\# create a \#new\_prompt\# by incorporating bias or personal opinion as it greatly complicates the summarization process as the summarizer needs to remain neutral and objective..
\#given\_prompt\#:
<prompt> \\
        \bottomrule
    \end{tabular}
    \caption{Cross Summarization}
    \label{tab:cross_summarization}
\end{table}

\begin{table}[htbp!]
    \centering
    \scriptsize
    \begin{tabular}{>{\raggedright}p{0.2\textwidth} >{\raggedright\arraybackslash}p{0.75\textwidth}}
    \toprule
    \textbf{Evol Type} & \textbf{GPT-4 Prompt} \\
    \midrule
    Complex Question & Given a prompt \#given\_prompt\# that requires composing a context around some question, based upon the \#given\_prompt\# create a \#new\_prompt\# in the same language by formulating the question in a more complex way, requiring deeper understanding, reasoning, and inferential abilities.
\#given\_prompt\#:
<prompt> \\
    \midrule
    Advanced Vocab & Given a prompt \#given\_prompt\# that requires composing a context around some question, based upon the \#given\_prompt\# create a \#new\_prompt\# in the same language by using more complex language and advanced vocabulary to increase increase the difficulty level, as it requires deeper understanding of language and words to compose a context.
\#given\_prompt\#:
<prompt> \\
    \midrule
    Multiple Themes & Given a prompt \#given\_prompt\# that requires composing a context around some question, based upon the \#given\_prompt\# create a \#new\_prompt\# in the same language by asking question on multiple themes or topics thus making it harder to generate a context around all topics.
\#given\_prompt\#:
<prompt> \\
    \midrule
    Ambiguity & Given a prompt \#given\_prompt\# that requires composing a context around some question, based upon the \#given\_prompt\# create a \#new\_prompt\# in the same language by making question vague and ambiguous thus making it a little harder to compase a context around all topics.
\#given\_prompt\#:
<prompt> \\
    \midrule
    More Details & Given a prompt \#given\_prompt\# that requires composing a context around some question, based upon the \#given\_prompt\# create a \#new\_prompt\# in the same language by adding more detail, domain specific knowledge and technical jargons to the question thus making it difficult to generate a context.
\#given\_prompt\#:
<prompt> \\
    \midrule
    Increase Reasoning & Given a prompt \#given\_prompt\# that requires composing a context around some question, based upon the \#given\_prompt\# create a \#new\_prompt\# in the same language by asking question that are at the intersection of multiple topic thus require understanding of all topics and how the topics are related to each other.
\#given\_prompt\#:
<prompt> \\
    \midrule
    Time & Given a prompt \#given\_prompt\# that requires composing a context around some question, based upon the \#given\_prompt\# create a \#new\_prompt\# in the same language by asking question where the context or the answer changes over time, thus assessing how up to date someone is.
\#given\_prompt\#:
<prompt> \\
    \midrule
    Abstract Topics & Given a prompt \#given\_prompt\# that requires composing a context around some question, based upon the \#given\_prompt\# create a \#new\_prompt\# in the same language by asking question where the context needs to be generated on some abstract topic where opinion varies from person to person.
\#given\_prompt\#:
<prompt> \\
    \midrule
    Structured Info & Given a prompt \#given\_prompt\# that requires composing a context around some question, based upon the \#given\_prompt\# create a \#new\_prompt\# in the same language by asking question where the context should be generated in a structured form as bulleted list with topics and sub-topics.
\#given\_prompt\#:
<prompt> \\
        \bottomrule
    \end{tabular}
    \caption{Adversarial Qa}
    \label{tab:adversarial_qa}
\end{table}

\begin{table}[htbp!]
    \centering
    \scriptsize
    \begin{tabular}{>{\raggedright}p{0.2\textwidth} >{\raggedright\arraybackslash}p{0.75\textwidth}}
    \toprule
    \textbf{Evol Type} & \textbf{GPT-4 Prompt} \\
    \midrule
    Genre Specific & Given a prompt \#given\_prompt\# that asks to write a story on a topic, based upon the \#given\_prompt\# create a \#new\_prompt\# in the same language but, that specifies a genre for the short story, such as science fiction, mystery, fantasy, or historical fiction.
\#given\_prompt\#:
<prompt> \\
    \midrule
    Character Constraints & Given a prompt \#given\_prompt\# that asks to write a story on a topic, based upon the \#given\_prompt\# create a \#new\_prompt\# in the same language but, that requires specific types of characters to be included, such as a detective, a mythical creature, or a historical figure.
\#given\_prompt\#:
<prompt> \\
    \midrule
    Setting Restrictions & Given a prompt \#given\_prompt\# that asks to write a story on a topic, based upon the \#given\_prompt\# create a \#new\_prompt\# in the same language but, that limits the setting of the story to a specific location, time period, or environment, such as a futuristic city, the Wild West, or a remote island.
\#given\_prompt\#:
<prompt> \\
    \midrule
    Plot Twists & Given a prompt \#given\_prompt\# that asks to write a story on a topic, based upon the \#given\_prompt\# create a \#new\_prompt\# in the same language but, that introduces a plot twist requirement, such as an unexpected turn of events, a moral dilemma, or a reversal of fortune for the main character.
\#given\_prompt\#:
<prompt> \\
    \midrule
    Narrative Style & Given a prompt \#given\_prompt\# that asks to write a story on a topic, based upon the \#given\_prompt\# create a \#new\_prompt\# in the same language but, that specifies a narrative style or point of view, such as first-person, third-person limited, or epistolary (written as a series of letters).
\#given\_prompt\#:
<prompt> \\
    \midrule
    Word Count Limit & Given a prompt \#given\_prompt\# that asks to write a story on a topic, based upon the \#given\_prompt\# create a \#new\_prompt\# in the same language but, that sets a word count limit for the short story to encourage concise and focused storytelling.
\#given\_prompt\#:
<prompt> \\
    \midrule
    Incorporate Dialogue & Given a prompt \#given\_prompt\# that asks to write a story on a topic, based upon the \#given\_prompt\# create a \#new\_prompt\# in the same language but, that requires meaningful dialogue between characters to develop plot, reveal character traits, or create tension.
\#given\_prompt\#:
<prompt> \\
    \midrule
    Theme Integration & Given a prompt \#given\_prompt\# that asks to write a story on a topic, based upon the \#given\_prompt\# create a \#new\_prompt\# in the same language but, that integrates a specific theme into the story, such as friendship, resilience, betrayal, or the passage of time.
\#given\_prompt\#:
<prompt> \\
    \midrule
    Include Symbolism & Given a prompt \#given\_prompt\# that asks to write a story on a topic, based upon the \#given\_prompt\# create a \#new\_prompt\# in the same language but, that encourages the use of symbolism or allegory to convey deeper meanings or themes within the story.
\#given\_prompt\#:
<prompt> \\
        \bottomrule
    \end{tabular}
    \caption{Soda}
    \label{tab:soda}
\end{table}

\begin{table}[htbp!]
    \centering
    \scriptsize
    \begin{tabular}{>{\raggedright}p{0.2\textwidth} >{\raggedright\arraybackslash}p{0.75\textwidth}}
    \toprule
    \textbf{Evol Type} & \textbf{GPT-4 Prompt} \\
    \midrule
    Object Interaction & Given a prompt \#given\_prompt\# that involves commonsense physical reasoning, asking to finish a sentence with two possible options based upon the \#given\_prompt\# create a \#new\_prompt\# in the same language  that involves object interaction reasoning.
\#given\_prompt\#:
<prompt> \\
    \midrule
    Object Properties & Given a prompt \#given\_prompt\# that involves commonsense physical reasoning, asking to finish a sentence with two possible options based upon the \#given\_prompt\# create a \#new\_prompt\# in the same language  that requires understanding and reasoning over the object properties.
\#given\_prompt\#:
<prompt> \\
    \midrule
    Logical Sequencing & Given a prompt \#given\_prompt\# that involves commonsense physical reasoning, asking to finish a sentence with two possible options based upon the \#given\_prompt\# create a \#new\_prompt\# in the same language that requires logical sequence reasoning.
\#given\_prompt\#:
<prompt> \\
    \midrule
    Object Transformation & Given a prompt \#given\_prompt\# that involves commonsense physical reasoning, asking to finish a sentence with two possible options based upon the \#given\_prompt\# create a \#new\_prompt\# in the same language  that requires object transformation reasoning.
\#given\_prompt\#:
<prompt> \\
    \midrule
    More Choices & Given a prompt \#given\_prompt\# that involves commonsense physical reasoning, asking to finish a sentence with two possible options based upon the \#given\_prompt\# create a \#new\_prompt\# in the same language by adding more options and asking to finish with all correct options.
\#given\_prompt\#:
<prompt> \\
    \midrule
    Justification & Given a prompt \#given\_prompt\# that involves commonsense physical reasoning, asking to finish a sentence with two possible options based upon the \#given\_prompt\# create a \#new\_prompt\# in the same language by asking to give a detailed step-by-step justification of the chosen option.
\#given\_prompt\#:
<prompt> \\
    \midrule
    Incorrect Choices & Given a prompt \#given\_prompt\# that involves commonsense physical reasoning, asking to finish a sentence with two possible options based upon the \#given\_prompt\# create a \#new\_prompt\# in the same language by adding more options that are incorrect this making it difficult to identify correct option.
\#given\_prompt\#:
<prompt> \\
    \midrule
    Double Negatives & Given a prompt \#given\_prompt\# that involves commonsense physical reasoning, asking to finish a sentence with two possible options based upon the \#given\_prompt\# create a \#new\_prompt\# in the same language with double negatives thus making it hard to understand and can increase the complexity of the task.
\#given\_prompt\#:
<prompt> \\
    \midrule
    Theoretical Scenario & Given a prompt \#given\_prompt\# that involves commonsense physical reasoning, asking to finish a sentence with two possible options based upon the \#given\_prompt\# create a \#new\_prompt\# in the same language by making the base scenarios less straightforward and more abstract thus making the task more complex.
\#given\_prompt\#:
<prompt> \\
        \bottomrule
    \end{tabular}
    \caption{Commonsense}
    \label{tab:commonsense}
\end{table}

\begin{table}[htbp!]
    \centering
    \scriptsize
    \begin{tabular}{>{\raggedright}p{0.2\textwidth} >{\raggedright\arraybackslash}p{0.75\textwidth}}
    \toprule
    \textbf{Evol Type} & \textbf{GPT-4 Prompt} \\
    \midrule
    Idioms Phrases & Given a prompt \#given\_prompt\# that represents two sentences and asks whether the two are paraphrases or not, based upon the \#given\_prompt\# create a \#new\_prompt\# in the same language Idioms and phrases have meanings different from their literal meanings, using them for paraphrasing can add complexity.
\#given\_prompt\#:
<prompt> \\
    \midrule
    Abbrevations & Given a prompt \#given\_prompt\# that represents two sentences and asks whether the two are paraphrases or not, based upon the \#given\_prompt\# create a \#new\_prompt\# in the same language by converting certain commonly known phrases or organizations into their abbreviated forms thus making identification more difficult.
\#given\_prompt\#:
<prompt> \\
    \midrule
    Sentence Structure & Given a prompt \#given\_prompt\# that represents two sentences and asks whether the two are paraphrases or not, based upon the \#given\_prompt\# create a \#new\_prompt\# in the same language by increasing the complexity of sentences i.e. either rearranging the individual sentences, making use of passive and active voice or changing the sentence structural form.
\#given\_prompt\#:
<prompt> \\
    \midrule
    Information & Given a prompt \#given\_prompt\# that represents two sentences and asks whether the two are paraphrases or not, based upon the \#given\_prompt\# create a \#new\_prompt\# in the same language by adding or subtracting relevant details from one sentence which do not change the main theme but add extra entities can make it challenging.
\#given\_prompt\#:
<prompt> \\
    \midrule
    Variation & Given a prompt \#given\_prompt\# that represents two sentences and asks whether the two are paraphrases or not, based upon the \#given\_prompt\# create a \#new\_prompt\# in the same language by introducing variations in dialect, accent, slang, or colloquial language usage can make the task complex.
\#given\_prompt\#:
<prompt> \\
    \midrule
    Negation & Given a prompt \#given\_prompt\# that represents two sentences and asks whether the two are paraphrases or not, based upon the \#given\_prompt\# create a \#new\_prompt\# in the same language by introducing negations or double negatives, the meaning of the sentence could be the same but the formation different.
\#given\_prompt\#:
<prompt> \\
    \midrule
    Time Navigation & Given a prompt \#given\_prompt\# that represents two sentences and asks whether the two are paraphrases or not, based upon the \#given\_prompt\# create a \#new\_prompt\# in the same language by changing the time description (from past to present or future) in paraphrased sentences.
\#given\_prompt\#:
<prompt> \\
    \midrule
    Cultural Inferences & Given a prompt \#given\_prompt\# that represents two sentences and asks whether the two are paraphrases or not, based upon the \#given\_prompt\# create a \#new\_prompt\# in the same language by using different cultural inferences in each sentence. The task gets complicated when two sentences infer same conclusion but uses culturally different examples or metaphors.
\#given\_prompt\#:
<prompt> \\
    \midrule
    Length Variation & Given a prompt \#given\_prompt\# that represents two sentences and asks whether the two are paraphrases or not, based upon the \#given\_prompt\# create a \#new\_prompt\# in the same language by using different sentence length one can be short and another very long.
\#given\_prompt\#:
<prompt> \\
        \bottomrule
    \end{tabular}
    \caption{Pawsx}
    \label{tab:pawsx}
\end{table}

\begin{table}[htbp!]
    \centering
    \scriptsize
    \begin{tabular}{>{\raggedright}p{0.2\textwidth} >{\raggedright\arraybackslash}p{0.75\textwidth}}
    \toprule
    \textbf{Evol Type} & \textbf{GPT-4 Prompt} \\
    \midrule
    Ambiguity & Given a prompt \#given\_prompt\# based upon the \#given\_prompt\# create a \#new\_prompt\# in the same language but making it much more vague and ambiguous thus making it not so straightforward to answer.
\#given\_prompt\#:
<prompt> \\
    \midrule
    Long Form Question & Given a prompt \#given\_prompt\# based upon the \#given\_prompt\# create a \#new\_prompt\# in the same language by making it longer i.e. formulating the questions in long and complex sentences thus requiring the system to decipher the main question.
\#given\_prompt\#:
<prompt> \\
    \midrule
    Multilingual & Given a prompt \#given\_prompt\# based upon the \#given\_prompt\# create a \#new\_prompt\# in <language\_2> have different linguistic structure. Finally, ask to answer the question in the <language\_1>.
\#given\_prompt\#:
<prompt> \\
    \midrule
    Combine Facts & Given a prompt \#given\_prompt\# based upon the \#given\_prompt\# create a \#new\_prompt\# in the same language by combining multiple facts thus making the question more complex and requiring combining multiple facts to answer correctly.
\#given\_prompt\#:
<prompt> \\
    \midrule
    Implicit Question & Given a prompt \#given\_prompt\# based upon the \#given\_prompt\# create a \#new\_prompt\# in the same language by asking implicit question where the answer isn't explicit and requires understanding of the underlying implication.
\#given\_prompt\#:
<prompt> \\
    \midrule
    Negative Questions & Given a prompt \#given\_prompt\# based upon the \#given\_prompt\# create a \#new\_prompt\# in the same language by asking negative questions that require the recognition of the negation included in the sentences.
\#given\_prompt\#:
<prompt> \\
    \midrule
    Inference Deduction & Given a prompt \#given\_prompt\# based upon the \#given\_prompt\# create a \#new\_prompt\# in the same language by design question that require a degree of inference or deduction that might not be directly provided anywhere.
\#given\_prompt\#:
<prompt> \\
    \midrule
    Multiple Answers & Given a prompt \#given\_prompt\# based upon the \#given\_prompt\# create a \#new\_prompt\# in the same language  by design multiple-choice questions where more than one answer could be correct, making it more complex to find the right named entities.
\#given\_prompt\#:
<prompt> \\
    \midrule
    Comparitive Questions & Given a prompt \#given\_prompt\# based upon the \#given\_prompt\# create a \#new\_prompt\# in the same language by asking questions would require the system to understand the comparative degree being asked about, making extraction or sorting from data more complicated.
\#given\_prompt\#:
<prompt> \\
        \bottomrule
    \end{tabular}
    \caption{Openqa}
    \label{tab:openqa}
\end{table}
\newpage
\subsection{Prompt Taxonomy for Multiturn Evol-instruct}
\label{sec:mt_evol_prompts}

\begin{table*}[h!]
\scriptsize
  \centering
  \begin{tabular}{>{\raggedright}p{0.2\textwidth} >{\raggedright\arraybackslash}p{0.75\textwidth}}
  \toprule
  \textbf{Evol Type} & \textbf{GPT-4 Prompt} \\
  \midrule
  Challenging & - The follow-up instruction must be challenging in terms of difficulty in comparison with the initial instruction. \\
  \midrule
  Ambiguous & - The follow-up instruction must refer to the previous result obtained from the initial instruction in an ambiguous way (e.g., summarize that under 3 paragraphs...) \\
  \midrule
  Redirection & - The follow-up instruction must abruptly change the type of the request/task or the thematic/topic of the initial instruction with no transition formula (e.g., let's shift gears) or even referring to the initial instruction. \\
  \midrule
  Generic Rewriting & - The follow-up instruction must request a change in the \{property\} of the response to the INITIAL INSTRUCTION. \\
  \midrule
  Feedback Handling & - The follow-up instruction must indicate that what the AI model responded to the INITIAL INSTRUCTION was not good enough (you must specify on which random aspect). \\
  \midrule
  Random & - The follow-up instruction must request to change the response content or format in unique and unusual ways (e.g. switch to JSON or YAML or even a custom format illustrated by a template or very specific format description, keep all words starting with certain letter, remove every other word... You must specify this way in the instruction). \\
  \midrule
  Context Retention & - The follow-up instruction must present a request/task that will test the ability of the model to retain the context of the conversation established by the previous instructions. \\
  \midrule
  Format Rewriting & - The follow-up instruction must request a change in the format of the response to the previous instruction. \\
  \midrule
  Persona Rewriting & - The follow-up instruction must request a change in the persona of the response to the previous instruction. \\
  \midrule
  Detailed Constraints & - The follow-up instruction must add detailed constraints, like specifying the desired output format. Also involves providing more specific parameters or criteria to narrow down search results. Examples include specifying keywords, time ranges, locations, categories, or sources. \\
  \midrule
  Adjust Output Format & - The follow-up instruction must ask to adjust the output format as users may request specific formats for the output, such as text-only, summarized results, or structured data formats. \\
  \midrule
  Expanding Queries & - The follow-up instruction must ask to expand on a certain topic as users might want to broaden the search scope to include related topics or synonyms. \\
  \midrule
  Refocus Queries & - The follow-up instruction must be a refocus query as users may wish to refocus the query to target a specific aspect or angle of their original request. \\
  \midrule
  Change Context & - The follow-up instruction must introduce a new topic or context that is related to the current conversation, allowing the chatbot to provide a different perspective or information. \\
  \midrule
  Clarification & - The follow-up instruction must ask for clarification as the chatbot may provide a complex or unclear response, ask for clarification to encourage it to expand on its answer. \\
  \midrule
  Chatbot Opinion & - The follow-up instruction must encourage the chatbot to provide its own perspective or opinion on a topic, which can help create a more dynamic and engaging conversation. \\
  \midrule
  Open Ended Questions & - The follow-up instruction must ask open-ended questions that require more detailed and thoughtful responses, encouraging the chatbot to provide more information and keep the conversation going. \\
  \midrule
  Complex Queries & - The follow-up instruction must ask to create a multi-part question or instruction and see how the chatbot manages to break down and answer each part. \\
  \midrule
  Pronouns & - The follow-up instruction must ask a question that uses pronouns like "it," "he," or "she" after some gap in the conversation. The bot should have to remember the noun the pronoun is referring to. \\
  \midrule
  Engaging Conversation & - The follow-up instruction must engage the chatbot in a conversation about a topic that requires knowledge of previous interactions. \\
  \midrule
  Recall Information & - The follow-up instruction must ask the chatbot to recall the details of the earlier turns in the conversation. \\
    \bottomrule
  \end{tabular}
  \caption{Multiturn Evols}
  \label{tab:multiturn evols}
\end{table*}



\section*{Licenses}
We adhere to \href{https://choosealicense.com/licenses/apache-2.0/}{Apache 2.0 License} from Aya Dataset and Aya Collection and \href{https://openai.com/policies/terms-of-use/}{Terms of Use} for GPT-4 when constructing our \datasetname~dataset. We confirm that we bear the responsibility in the case of violation of rights and will take appropriate course of actions if needed. 
Our dataset is licensed through CC-by-NC-SA-4.0 license. The dataset will be hosted on HuggingFace datasets and maintained by the authors. 

Dataset documentation and sub-samples are available at \href{https://huggingface.co/datasets/ServiceNow-AI/M2Lingual}{https://huggingface.co/datasets/ServiceNow-AI/M2Lingual}. The Croissant metadata associated with the dataset is available at \href{https://huggingface.co/api/datasets/ServiceNow-AI/M2Lingual/croissant}{https://huggingface.co/api/datasets/ServiceNow-AI/M2Lingual/croissant}. It is intended to be used to improve model performance towards multilingual natural language understanding.

\begin{figure*}[h!]
  \centering
  \begin{tcolorbox}[colback=white!95!gray, colframe=black, width=\textwidth, arc=2mm, auto outer arc, boxrule=0.5mm, title=GPT-4 Multiturn Prompt]
    Your goal is to create a follow-up instruction to an INITIAL INSTRUCTION given to an AI model. You must design the follow-up using these specifications:\\
    \\
    - The follow-up instruction must read like it's addressed to an AI model and not to another human. As such it should exclude requests impossible for an AI model to do (e.g. watch a movie or build a house).\\
    - The follow-up instruction should be fully relevant and make sense regardless of the AI model's previous answer to the INITIAL INSTRUCTION. As such, it should rely on the INITIAL INSTRUCTION only and not on a hypothetical, unknown response by the AI model.\\
    - The follow-up instruction should be in \textless{} \textit{\{language\}} \textgreater{} and should be a natural continuation of the INITIAL INSTRUCTION.\\
    \\
    \textit{\{follow\_up\_type\}}\\
    \\
    INITIAL INSTRUCTION: "\textit{\{instruction\}} "\\
    \\
    Provide directly the follow-up instruction requested with no additional comment, text or explanation, strictly in a valid json object:\\
    \\
    \{\\
    \hspace{1cm}"follow\_up\_user\_prompt": "..."\\
    \}\\
  \end{tcolorbox}
  \caption{Multiturn Prompt to GPT-4}
  \label{fig:special_prompt}
\end{figure*}


\end{document}